%% file: main.tex
\documentclass[10pt,twocolumn,letterpaper]{article}

\usepackage[pagenumbers]{cvpr} 

\input{preamble}

\definecolor{cvprblue}{rgb}{0.21,0.49,0.74}
\usepackage[pagebackref,breaklinks,colorlinks,allcolors=cvprblue]{hyperref}
\usepackage{verbatim}

\usepackage{wrapfig}  
\usepackage{graphicx}
\usepackage{floatrow}
\usepackage{subcaption}
\usepackage{listings}
\usepackage{algorithm}

\def\confName{CVPR}
\def\confYear{2025}

\title{Causal Diffusion Transformers for Generative Modeling}

\author{Chaorui Deng ~~~~ Deyao Zhu ~~~~ Kunchang Li ~~~~ Shi Guang ~~~~ Haoqi Fan\\
ByteDance Research\\
\tt\small{\href{https://github.com/causalfusion/causalfusion.git}{https://github.com/causalfusion}} \\
}

\begin{document}
\maketitle
\input{sec/0_abstract}
\input{sec/1_intro}
\input{sec/2_related}
\input{sec/3_method}
\input{sec/5_conclusion}

\input{sec/appendix}

\clearpage
{
    \small
    \bibliographystyle{ieeenat_fullname}
    \bibliography{main}
}


\end{document}

%% file: preamble.tex
\usepackage{xcolor}
\usepackage{graphicx}
\usepackage{booktabs}
\usepackage{amsmath} 
\usepackage{amsfonts}
\usepackage{amssymb}
\usepackage{multirow} 
\usepackage{makecell}
\newcommand{\shline}{\Xhline{1.1pt}} 

%% file: sec/0_abstract.tex
\begin{abstract}
We introduce Causal Diffusion as the autoregressive (AR) counterpart of Diffusion models. It is a next-token(s) forecasting framework that is friendly to both discrete and continuous modalities and compatible with existing next-token prediction models like LLaMA and GPT. While recent works attempt to combine diffusion with AR models, we show that introducing sequential factorization to a diffusion model can substantially improve its performance and enables a smooth transition between AR and diffusion generation modes. Hence, we propose \textbf{CausalFusion} - a decoder-only transformer that dual-factorizes data across sequential tokens and diffusion noise levels, leading to state-of-the-art results on the ImageNet generation benchmark while also enjoying the AR advantage of generating an arbitrary number of tokens for in-context reasoning. We further demonstrate CausalFusion's multimodal capabilities through a joint image generation and captioning model, and showcase CausalFusion's ability for zero-shot in-context image manipulations. We hope that this work could provide the community with a fresh perspective on training multimodal models over discrete and continuous data.
\end{abstract}

%% file: sec/1_intro.tex
\vspace{-10pt}
\section{Introduction}
\label{sec:intro}
Autoregressive (AR) and diffusion models are two powerful paradigms for data distribution modeling. AR models, also known as the next token prediction approach, dominate language modeling and are considered central to the success of large language models (LLMs)~\cite{gpt1,gpt2,gpt3,llama1,llama2,llama3}. On the other hand, diffusion models~\cite{ddpm,dit,adm,edm}, or score-based generative models~\cite{songscore,lipman2023flow}, have emerged as the leading approach for visual generation, driving unprecedented progress in the era of visual content generation~\cite{sora,rombach2022high,li2023scaling}. 

\begin{figure}[t]
    \centering
    \includegraphics[width=1.0\textwidth,height=1.0\textwidth]{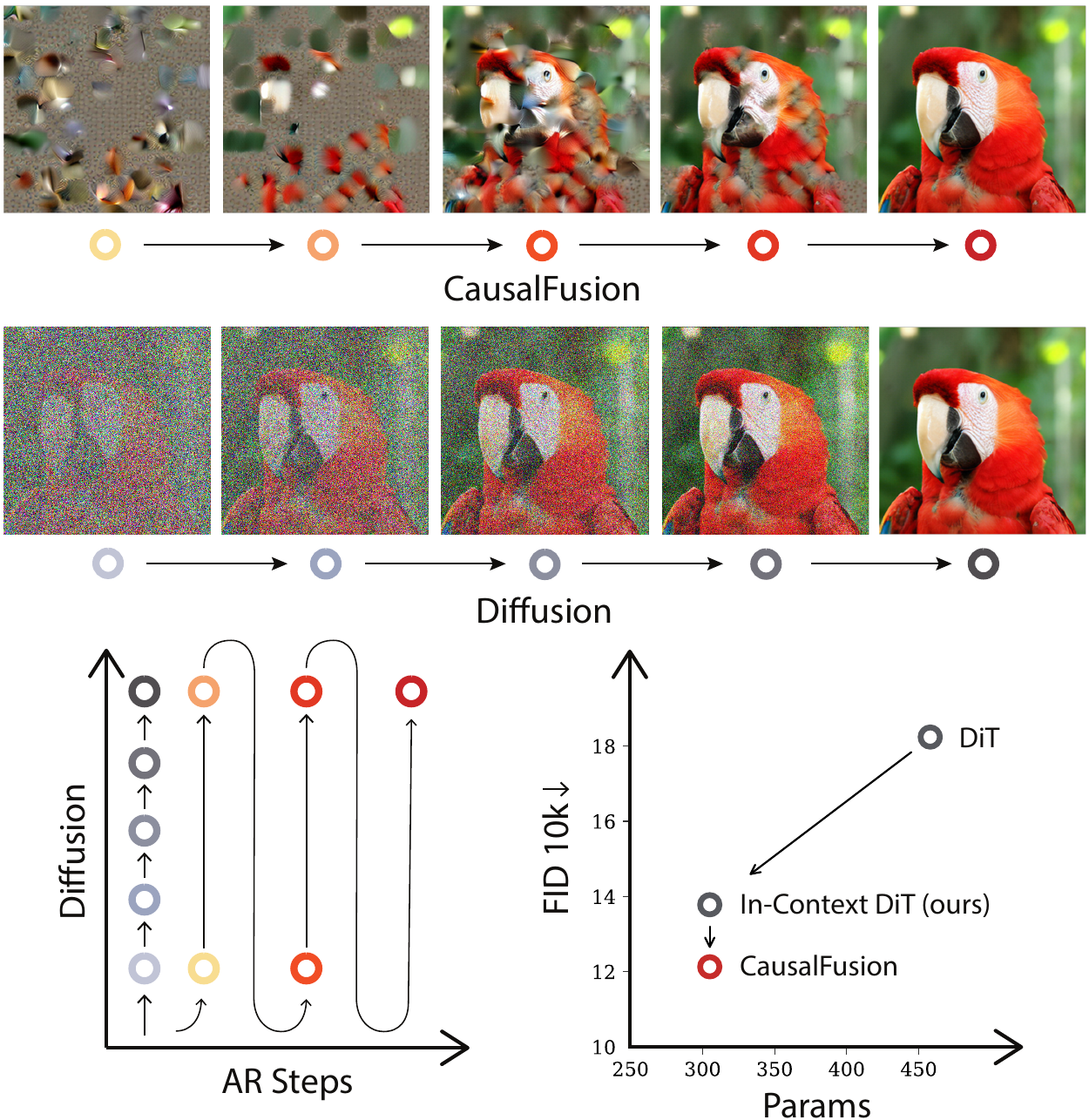} 
    \vspace*{-6mm}
    \caption{
    \textbf{Illustration of Dual-Factorization}. The arrow line indicates CausalFusion's generation path, moving from one state to the next by jointly generating along the sequential and noise-level dimension at each step. 
    Compared to DiT, our In-context DiT substantially improves results with fewer parameters. CausalFusion further enhances performance without changing the architecture or parameter count. Results were trained on IN1K for 240 epochs. CausalFusion adopts arbitrary AR steps for image generation, but each step only diffuses partial tokens, resulting in similar (or slightly lower) computational complexity.
    \vspace{-10pt}
    }
    \label{fig:dual-factorization}
\end{figure}

\begin{figure*}[t]
  \centering
  \begin{subfigure}{1.0\linewidth}
    \centering
    \includegraphics[width=\linewidth]{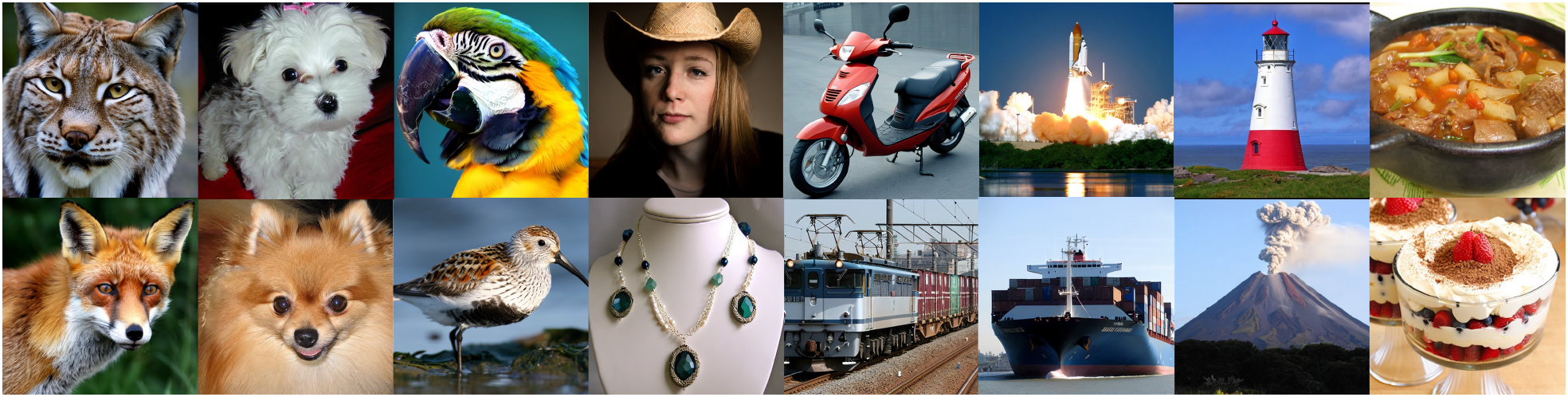}
    \caption{Samples generated by CausalFusion-XL/2, ImageNet 512$\times$512, 800 epoch, DDPM 250 steps, CFG=4.0}
  \end{subfigure}
  \begin{subfigure}{1.0\linewidth}
    \centering
    \includegraphics[width=\linewidth]{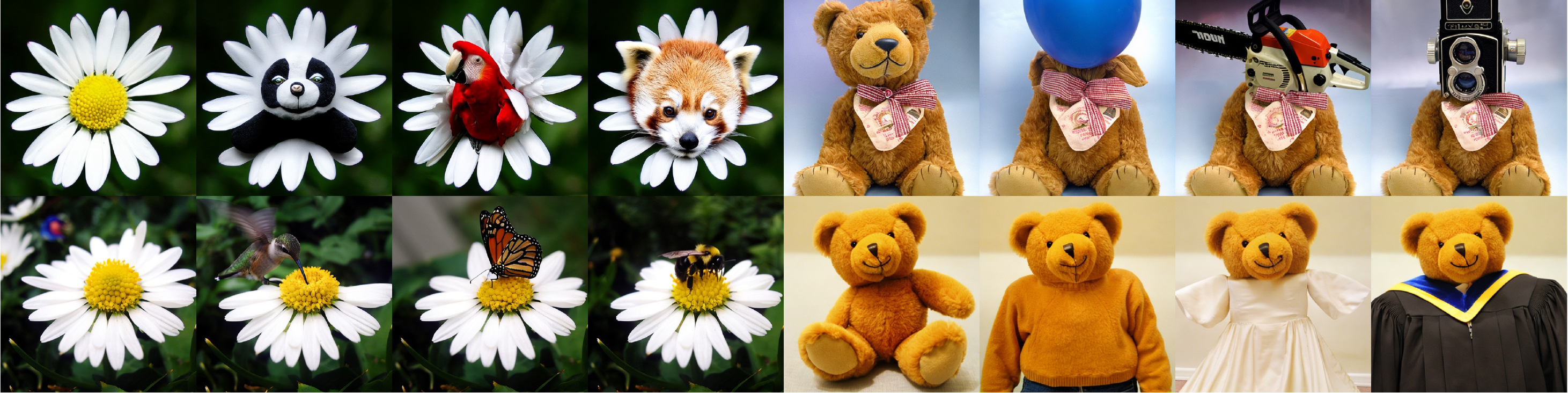}
    \caption{\textbf{Zero-shot image editing} results generated by CausalFusion-XL/2, ImageNet 512$\times$512, 800 epoch. We first generate the original image (those on the left), then mask out its centre region, top-half, or bottom-half, and regenerate the image with new class conditions. Details are discussed in Sec \ref{sec:system}.}
  \end{subfigure}
  \caption{\textbf{Visualization results}. All samples are generated by models trained only on \textbf{ImageNet-1K class-conditional generation} task, demonstrating CausalFusion's zero-shot image manipulation ability. See more visualization results in Appendix~\ref{appendix:secD}.
  \vspace{-12pt}
  }
  \vspace{-6pt}
  \label{fig:vis1}
\end{figure*}

The intrinsic distinction between AR and diffusion models lies in their approach to data distribution factorization. AR models treat data as an ordered sequence, factorizing it along the sequential axis, where the probability of each token is conditioned on all preceding tokens. This factorization enables the AR paradigm to generalize effectively and efficiently across arbitrary number of tokens, making it well-suited for long-sequence reasoning and in-context generation. In contrast, diffusion models factorize data along the noise-level axis, where the tokens at each step are a refined (denoised) version of themselves from the previous step. As a result, the diffusion paradigm is generalizable to arbitrary number of data refinement steps, enabling iterative quality improvement with scaled inference compute. While AR and diffusion models each excel within their respective domains, their distinct factorization approaches reveal complementary potential. Although recent studies~\cite{transfusion,monoformer,dart} have attempted to integrate AR and diffusion within a single model, they typically treat these paradigms as separate modes, missing the potential benefits of jointly exploring them within a 2-D factorization plane.

To this end, we introduce \textbf{CausalFusion}, a flexible framework that integrates both sequential and noise-level data factorization to unify their advantages. The degree of factorization along these two axes—namely, the AR step and diffusion step—is adjustable, enabling {CausalFusion} to revert seamlessly to the traditional AR or diffusion paradigms at either extreme. To enhance its generality, CausalFusion is designed to predict \textit{any} number of tokens at \textit{any} AR step, with \textit{any} pre-defined sequence order and \textit{any} level of inference compute, thereby minimizing the inductive biases presented in existing generative models. As shown in Figure~\ref{fig:dual-factorization}, this approach provides a broad spectrum between the AR and diffusion paradigms, allowing smooth interpolation within two endpoints during both training and inference. 
Specifically, we explore CausalFusion in image generation and multimodal generation scenarios, where we observe that the level of training difficulties significantly influences the overall effectiveness of CausalFusion.

\textbf{Difficulties of generative tasks in CausalFusion:} Both AR and diffusion paradigms present unique challenges based on difficulties of their specific generative stages. In diffusion models, the effectiveness of training depends heavily on proper loss weighting across noise levels~\cite{ddpm,minsnr}, as higher noise levels are more difficult and usually provide more valuable signals than lower noise levels. Similarly, AR models are susceptible to error accumulation~\cite{bengio2015scheduled} as early-stage predictions are made with limited visible context, making them more error-prone. Optimizing CausalFusion thus requires balancing across these varying task difficulties to optimize training signal impact and ensure sufficient exploration across the entire factorization plane.

In this paper, we formally examine the difficulties of generative tasks within CausalFusion. We show that, in addition to the noise levels in diffusion and the amount of visible context in AR, the total number of AR steps, which controls the interpolation between AR and diffusion, also plays a critical role in shaping training difficulties. Driven by these factors, we develop a scalable and versatile model based on the CausalFusion framework. Starting from the DiT architecture~\cite{dit}, we gradually convert it into a decoder-only transformer compatible with existing AR models like GPT~\cite{gpt1,gpt2,gpt3} and LLaMA~\cite{llama1,llama2,llama3}. We provide insights on how to appropriately choose the number of AR steps during the training of CausalFusion models, and further introduce loss weighing along both the diffusion and AR axis to balance the impact of different generative stages. As shown in Figure~\ref{fig:dual-factorization} and ~\ref{fig:vis1}, our model achieves state-of-the-art performance on the ImageNet class-conditional generation benchmark, significantly outperforming DiT~\cite{dit} and enabling zero-shot image manipulations due to its AR nature. When pretraining on both text-to-image and image-to-text tasks, our model surpasses forced-fusion frameworks such as TransFusion~\cite{transfusion}, demonstrating the versatility of our CausalFusion framework.

We highlight our main contribution below:
\begin{itemize}
\item  We propose CausalFusion as the AR counterpart to DiT, achieving state-of-the-art results and enabling the unlimited token generation for in-context reasoning.
\item  We systematically study CausalFusion on the dual-factorization plane and identify key factors that improve the effectiveness of CausalFusion models.
\item  Compared with recent studies~\cite{transfusion}, CausalFusion enables a smooth, cohesive integration with language modeling for cross-modal generation and reasoning.
\end{itemize} 

%% file: sec/2_related.tex
\section{Related Works}
\label{sec:related}

\textbf{Diffusion Models}.
Diffusion models \cite{sohl2015deep, song0, ddpm} decompose the image generation task into a sequence of iterative denoising steps, gradually transforming noise into a coherent image. Early diffusion models \cite{adm, rombach2022high, sdxl, nichol2021glide, dalle2} with U-net architectures pioneered denoising techniques for high-quality image synthesis. Later works \cite{dit, bao2023all} like DiT shift from the U-net to transformer-based architectures, enabling greater compute scalability. Modern methods \cite{, chen2024pixart, flux} further extend DiT architectures leveraging significantly larger training resources to achieve impressive image generation quality.

\vspace{3pt}
\textbf{Autoregressive Generation}.
Another popular approach for image generation involves autoregressive (AR) transformers that predict images token by token. Early works \cite{dalle1, ding2021cogview, gafni2022make, parti} generated image tokens in raster order, progressing sequentially across the image grid. This rasterized approach was later identified as inefficient \cite{chang2022maskgit}, prompting researchers to explore random-order generation methods \cite{chang2022maskgit, muse}. 
AR methods are further evolved to include new modalities such as video generation~\cite{kondratyukvideopoet} and any-to-any generation~\cite{io2, anygpt}. 

\vspace{3pt}
\textbf{Combining Diffusion and Autoregressive Models}.
Recent models explore different methods for integrating AR and diffusion processes. DART \cite{dart} unifies AR and diffusion in a non-Markovian framework by conditioning on multiple historical denoising steps instead of only the current one. BiGR \cite{bigr} generates discrete binary image codes autoregressively using a Bernoulli diffusion process. MAR \cite{mar} employs an AR model with a small diffusion head to enable continuous-value generation. Emu2 \cite{emu2} applies an external diffusion module to decode its AR-based multimodal outputs. 
Compared to previous methods, CausalFusion focuses on autoregressive sequence factorization and decouples diffusion data processing across both sequential tokens and noise levels, achieving significant performance gains over traditional diffusion frameworks.

%% file: sec/3_method.tex
\section{CausalFusion}

\paragraph{Preliminaries.}\label{sec:background}
We first briefly review the Autoregressive (AR) and Diffusion paradigms in the context of image modeling before introducing our CausalFusion model. Both paradigms factorize the image distribution into a chain of conditional distributions. However, they do so along different axes.
Given a sample of training images $\mathbf{X}$, AR models split $ \mathbf{X} $ along the spatial dimensions into a sequence of tokens, $ \mathbf{X} = \{\mathbf{x}_{1}, \dots, \mathbf{x}_{L}\} $, where $ L $ is the number of tokens. The joint distribution of $ \mathbf{X} $ can be then factorized sequentially:
\begin{equation}\label{eq:AR-factorization}
 q(\mathbf{x}_{1:L}) = q(\mathbf{x}_{1}) \prod_{l=2}^{L} q(\mathbf{x}_{l} | \mathbf{x}_{1:l-1}).
\end{equation}
During training, a neural network $ p_\theta(\mathbf{x}_{l} | \mathbf{x}_{1:l-1}) $ is trained to approximate $ q(\mathbf{x}_{l} | \mathbf{x}_{1:l-1}) $ by minimizing the cross-entropy $ -\mathbb{E}_{q(\mathbf{x}_{1:L})} \log p_\theta(\mathbf{x}_{1:L}) $. At inference time, the image is generated via the \textit{next token prediction} paradigm.

In contrast, Diffusion models gradually add random noise (typically Gaussian) to $\mathbf{X}$ in a so-called \textit{forward process}. It is a Markov chain along the noise level, where each noisy version $\mathbf{x}_t$ is conditioned on the previous state $ \mathbf{x}_{t-1} $ as $ q(\mathbf{x}_t | \mathbf{x}_{t-1}) = \mathcal{N}(\mathbf{x}_t; \sqrt{1 - \beta_t} \mathbf{x}_{t-1}, \beta_t \mathbf{I}) $. Here, $ \beta_t $ is a variance schedule that ensures the forward process starts with a clean image $ \mathbf{x}_0 = \mathbf{X} $ and gradually converges to random noise as $ t \rightarrow T $. The joint distribution of $ \mathbf{X} $ is then factorized as: 
\begin{equation}\label{eq:diffusion-factorization}
q(\mathbf{x}_{0:T}) = q(\mathbf{x}_0)\prod_{t=1}^T q(\mathbf{x}_t | \mathbf{x}_{t-1}).
\end{equation}
To obtain $ \mathbf{X} $ from noise, a neural network is trained to approximate the reverse transition of the forward process for $ t \in [1, T] $:
\begin{equation}\label{eq:diffusion-reverse}
p_\theta(\mathbf{x}_{t-1} | \mathbf{x}_t) = \mathcal{N}(\mathbf{x}_{t-1}; {\mu_\theta}(\mathbf{x}_t), \Sigma_\theta(\mathbf{x}_t))
\end{equation}
As in AR models, training involves minimizing the cross-entropy between $ q(\mathbf{x}_{0:T}) $ and $p_\theta(\mathbf{x}_{0:T})$. 
In DDPM~\cite{ddpm}, $\Sigma_\theta(\mathbf{x}_t)$ is set to a constant value derived from the forward process, and $\mu_\theta(\mathbf{x}_t)$ is set to be the linear combination of $\mathbf{x}_t$ and a noise prediction model $\epsilon_\theta$ that predicts the noise $\mathbf{\epsilon}$ of the forward process. This parameterization leads to the following training objective:
\begin{equation}\label{eq:diffusion_loss}
    \min_\theta \mathbb{E}_{\mathbf{x}_0, \mathbf{\epsilon}, t}[w(t) \|\mathbf{\epsilon} - \epsilon_\theta(\mathbf{x}_t, t)\|^2] 
\end{equation}
where $w(t)$ is derived according to the noise schedule $\beta_t$, which gradually decays as $t$ grows. This objective is further simplified by setting $w(t) = 1$ for all $t$, results in a weighted evidence lower bound that emphasizes more difficult denoising tasks (i.e., larger noise level) at larger $t$ step.

\begin{figure}[t]
    \centering
    \includegraphics[width=1.02\linewidth]{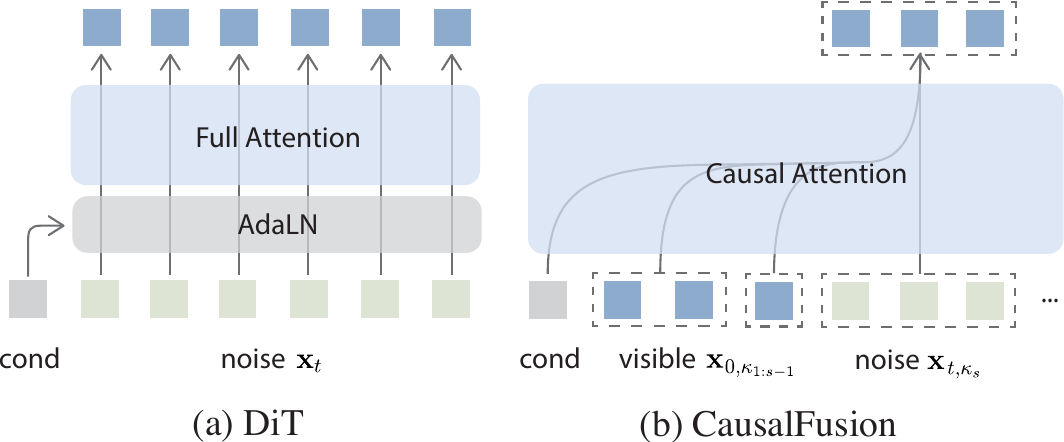}
    \caption{\textbf{Conceptual comparison between the DiT and CausalFusion architectures}. a) DiT incorporates conditioning via adaptive layer normalization, processing a fixed-size set of entire image tokens as input. All the noise tokens $x_t$ are fed into DiT with full attention observation, enabling comprehensive modeling of the input during processing.
    b) CausalFusion treats all input modalities equally in an in-context manner, denoising a random subset of image tokens $x_{t, \kappa_s}$ at each step while causally conditioning on previously denoised tokens $x_{0, 1:\kappa_{s-1}}$, and other contextual inputs. This approach enforces the model to reconstruct the image with partial observation, embodying the spirit of masked feature prediction models \cite{he2022masked, maskedpredict, flip}.
    \vspace{-5pt}
    }
    \label{fig:arch}
\end{figure}

\paragraph{Our approach.}
From the above formulations, the AR and diffusion paradigms naturally support the scaling of sequence length and denoising steps, respectively, offering complementary advantages for image generation. To unify their advantages, we present CausalFusion, a general paradigm that scales effectively in both directions. 

We start by directly extending Eqn.~(\ref{eq:diffusion-factorization}) to encompass the AR factorization:
\begin{align}\label{eq:cd-factorization}
& q(\mathbf{x}_{0:T,\kappa_s} | \mathbf{x}_{0,\kappa_{1:s-1}}) = \nonumber \\ 
& q(\mathbf{x}_{0,\kappa_s}) \prod_{t=1}^T q(\mathbf{x}_{t,\kappa_s} | \mathbf{x}_{t-1,\kappa_s},\mathbf{x}_{0,\kappa_{1:s-1}}) \tag{5}
\end{align}
for $s \in [1, S]$. Here, $S$ denotes the total number of AR steps, while $\kappa_s$ is an index set that identifies the subset of image tokens to be processed at the $s$-th AR step, with $|\kappa_s|$ representing the number of tokens in this subset. Each AR step processes only the tokens indicated by $\kappa_s$, isolating specific portions of the image, as shown in the top row of Figure~\ref{fig:dual-factorization}. The term $\mathbf{x}_{t,\kappa_s}$ represents the dual-factorized image tokens at the $s$-th AR step and $t$-th diffusion step.

During training, the objective of our CausalFusion model is to approximate $p_\theta(\mathbf{x}_{t-1,\kappa_s} | \mathbf{x}_{t,\kappa_s},\mathbf{x}_{0,\kappa_{1:s-1}})$ for all $t$ and $s$. Compared with the formulation in Eqn.~(\ref{eq:diffusion-reverse}),
CausalFusion requires the training sequence to contain not only noised image tokens at the current AR step $\mathbf{x}_{t,\kappa_s}$, but also the clean image tokens from all previous AR steps $\mathbf{x}_{0,\kappa_{1:s-1}}$, allowing the model to leverage information from earlier AR steps to refine the current tokens effectively.
A generalized version of causal attention mask is also required to prevent $\mathbf{x}_{0,\kappa_{1:s-1}}$ from observing $\mathbf{x}_{t,\kappa_s}$.
During inference, the dual-factroization in Eqn.~(\ref{eq:cd-factorization}) enables CausalFusion to generate an unlimited sequence of image tokens through a \textit{next token(s) diffusion} approach while enhancing the quality of each token by applying larger numbers of diffusion steps. See Figure~\ref{fig:arch}(b) for an illustration of CausalFusion model architecture. Further details on the generalized causal attention mask can be found in Appendix~\ref{appendix:secA}.

Notably, adhering to the principle of \textbf{minimal inductive bias}, CausalFusion imposes no restrictions on the number of AR steps $S$, the number of tokens processed at each AR step $|\kappa_s|$, or the specific token indices within each AR step. This flexibility enables a broad exploration space for generative modeling in both training and inference stages.

\begin{table}[t]
\footnotesize
\centering
\begin{tabular}{l|cc}
Model & Params (M) & FID10k$\downarrow$ \\ 
\shline
\underline{DiT}~\cite{dit} & 458 & {18.24}  \\
\hline
{- AdaLN-zero}~\cite{dit} & {305} & {26.71} \\
~~+ new recipe & 305 & {21.94}   \\
~~~~+ T embedding & 308 & {20.68}  \\
~~~~~~+ QK-norm & 308 & {18.66} \\
~~~~~~+ lr warmup & 308 & 17.11 \\
\hline
\colorbox{gray!30}{+ All} (In-context DiT) & 308 &  \textbf{13.78}
\end{tabular}
\caption{\textbf{In-context DiT baseline}. ImageNet 256$\times$256, 240 epoch. Baseline settings are marked by \underline{underlines} and selected settings highlighted in \colorbox{gray!30}{gray}.
\vspace{-6pt}
}
\label{tab:in-context-dit}
\end{table}

\section{Initial studies on CausalFusion}\label{sec:init_exps}
To systematically study the design space of CausalFusion, we conduct experiments on the ImageNet dataset~\cite{deng2009imagenet}, where we train class-conditional image generation models at 256$\times$256 resolution. We use the DiT-L/2 model as our base configuration. All models are trained with 240 epochs and evaluated using FID-10k (unless specified, we use FID-10K and FID interchangeably) and the ADM~\cite{adm} codebase. 
Detailed training recipes and model configurations are provided in Appendix~\ref{appendix:secC}.

\begin{figure*}[t]
  \centering
    \begin{subfigure}{0.32\textwidth}
        \centering
        \includegraphics[width=\linewidth]{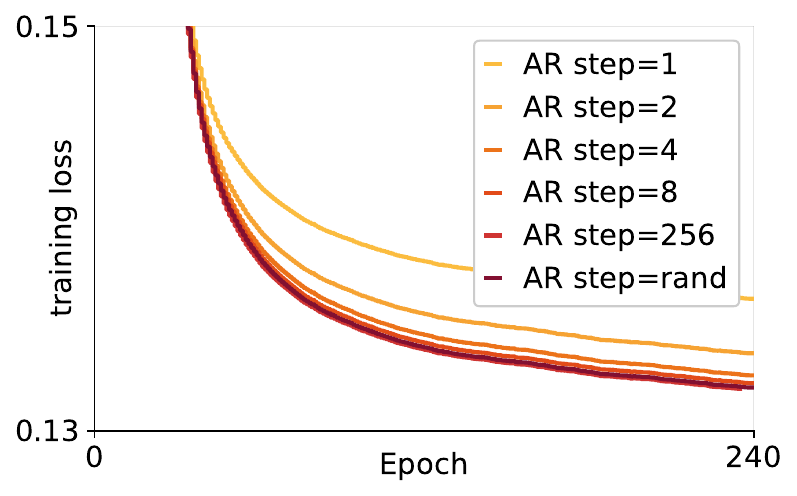}
        \caption{}
        \label{fig:fig1}
    \end{subfigure}
    \hfill
    \begin{subfigure}{0.32\textwidth}
        \centering
        \includegraphics[width=\linewidth]{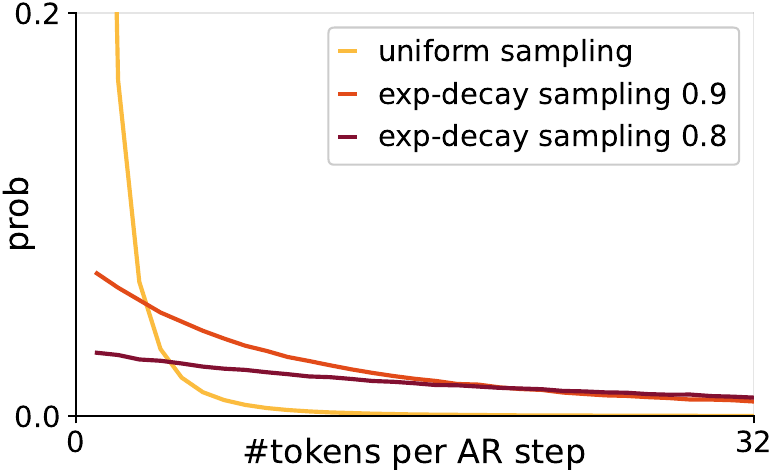}
        \caption{}
        \label{fig:fig2}
    \end{subfigure}
    \hfill
    \begin{subfigure}{0.32\textwidth}
        \centering
        \includegraphics[width=\linewidth]{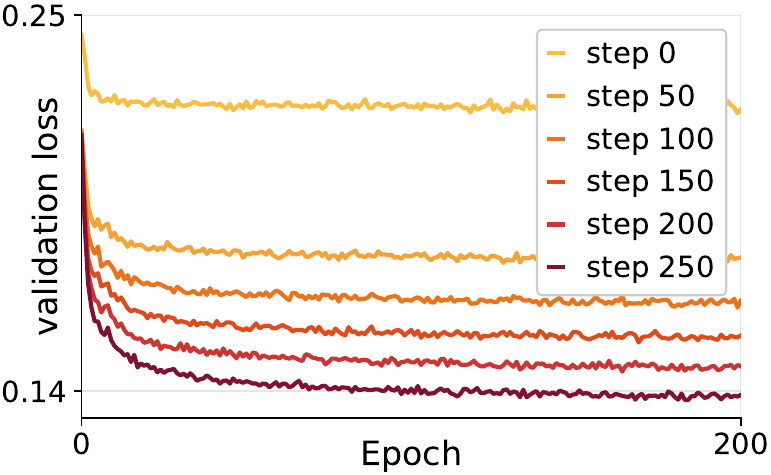}
        \caption{}
        \label{fig:fig3}
    \end{subfigure}
   \vspace{-5pt}
   \caption{
   (a) Training loss using different number of AR steps. (b) Distribution of $|\kappa_s|$. (c) Validation loss at difference AR steps. 
   \vspace{-6pt}
   }
   \label{fig:weighting_and_decay}
\end{figure*}

\paragraph{Baseline setup: In-context DiT.} As our target is to unify the AR and Diffusion paradigms, we need to unify their architectures first. To this end, we begin with the Transformer-based DiT~\cite{dit} model. Following the DiT design, the 256$\times$256 images are encoded into 32$\times$32 latent representations~\cite{rombach2022high} using a pretrained VAE model, followed by a 2$\times$2 patchify layer that produces a sequence of $L$ = 256 latent tokens. In original DiT, conditional information (e.g., label class) and the diffusion time step are incorporated through AdaLN-zero components, which are incompatible with decoder-only LLMs. To address this limitation, we adopt the in-context design of DiT from~\cite{dit}, treating the class and time step conditions as tokens and directly append them to the image token sequence. By default, we use 4 class tokens and one time step token. As a byproduct, this modification reduces the model size of in-context DiT to approximately $\frac{2}{3}$ of the AdaLN-zero version.

To accelerate training, we use large batch sizes (e.g., 2048) and implement several improvements to stabilize training: (1) injecting the diffusion time step by adding a time step embedding to the image token embeddings instead of using a time step token; (2) applying head-wise QK layer normalization within the self-attention layers, following the practices in~\cite{dehghani2023vit22b}; and (3) incorporating a learning rate warmup stage during training. 

The impact of our new designs is shown in Table~\ref{tab:in-context-dit}. Initially, the native In-context DiT from~\cite{dit} performs significantly worse than the AdaLN-zero version. Our revised training recipe improves performance to 21.94 FID. Incorporating the time step embedding and head-wise QK norm further enhances performance, achieving an FID of 18.66. Adding the learning rate warmup yields an additional improvement. Overall, our final in-context DiT-L/2 model, though conceptually simple, reaches an FID-10k of 13.78—competitive to the best-performing DiT-XL/2 model (12.92 FID-10k) in~\cite{dit}—and serves as a robust baseline that can be steadily trained with large batch sizes.

\begin{table}[t]
\footnotesize
\centering
\begin{tabular}{c|cccc}
 & \multicolumn{4}{c}{FID10k$\downarrow$} \\ 
\#AR steps  & $S_\text{eval}$ = 1 & $S_\text{eval}$ = 2 & $S_\text{eval}$ = 4 & $S_\text{eval}$ = 8\\ 
\shline
\underline{$S_\text{train}$ = 1} & \textbf{13.78} & 356.69 & 404.67 & 390.18 \\ 
$S_\text{train}$ = 2  & 16.69 & \textbf{14.77} & 47.49 & 136.04 \\
$S_\text{train}$ = 4  & 24.14 & 15.37 & \textbf{18.13} & 33.14\\
$S_\text{train}$ = 8  & 54.08 & 24.49 & 22.66 & \textbf{20.01}  \\
$S_\text{train}$ = 256 & 313.28 & 321.62 & 261.26 & 192.25 \\
\hline
\colorbox{gray!30}{random} & 21.31 & 22.17 & 23.54 & 25.05
\end{tabular}
\caption{\textbf{Ablations on AR steps}. $S_\text{train}$ and $S_\text{eval}$ indicates the fixed AR steps used during training and inference, respectively. Baseline settings are marked by \underline{underlines} and selected settings highlighted in \colorbox{gray!30}{gray}.
\vspace{-10pt}
}
\label{tab:ng}
\end{table}

\begin{table*}[t]
\footnotesize
\centering
\subfloat[\textbf{Exponential decay} in AR step sampling. A proper decay ratio leads to competitive or better performance across all inference settings than using fixed AR steps.]{
\begin{tabular}{c|cccc}
 & \multicolumn{4}{c}{FID10k$\downarrow$} \\ 
 {ratio}   & $S_\text{eval}$ = 1 & $S_\text{eval}$ = 2 & $S_\text{eval}$ = 4 & $S_\text{eval}$ = 8 \\ 
\shline
\underline{1.0} & 21.31 & 22.17 & 23.54 & 25.05\\
0.95 & {14.49} & 17.78 & 19.79 & 23.93\\
\colorbox{gray!30}{0.9} &  {12.89} & 15.57 & \textbf{18.83} & \textbf{22.72} \\
0.85 & 12.94 & 15.54 & 19.12 & 23.46\\
0.8 & \textbf{12.78} & \textbf{15.42} & 19.38 & 23.78\\
\end{tabular}
}
\hfill
\subfloat[\textbf{Token order} influences the locality of image tokens and further affects training difficulty.]{
\begin{tabular}{c|c}
Patch order & FID10k$\downarrow$ \\ 
\shline
raster order & {14.46} \\
block-wise raster (8x8) & 14.76 \\
block-wise raster (4x4) & 14.62 \\
dilated order & 15.54 \\
\colorbox{gray!30}{\underline{random order}} & {\textbf{12.89}} \\
\end{tabular}
}
\hfill
\subfloat[\textbf{AR loss weighting} boosts performance by facilitating better learning from difficult samples.]{
\footnotesize
\centering
\begin{tabular}{c|ccc}
 & \multicolumn{3}{c}{FID10k$\downarrow$} \\ 
weight & $S_\text{eval}$ = 1 & $S_\text{eval}$ = 2 & $S_\text{eval}$ = 4 \\ 
\shline
\underline{1$\rightarrow$1} & {12.89} & 15.57 & 18.83 \\
1.5$\rightarrow$1 & {12.61} & 15.49 &  18.32 \\
\colorbox{gray!30}{2$\rightarrow$1} & {\textbf{12.13}} & \textbf{15.15} & 18.09\\
2.5$\rightarrow$1 & {12.32} & 15.22 & 17.99 \\
3$\rightarrow$1 & {12.50} & 15.28 & \textbf{17.92}\\
\end{tabular}
}
\vspace{-5pt}
\caption{\textbf{Ablations on AR step decay, ordering, and AR weighting}. Baseline settings are marked by \underline{underlines} and selected settings highlighted in \colorbox{gray!30}{gray}.
\vspace{-10pt}
}
\label{tab:arweighting}
\end{table*}

\vspace{-5pt}
\paragraph{CausalFusion with fixed number of AR steps.} Building on the In-context DiT baseline, we begin with a simplified version of CausalFusion that uses a fixed number of AR steps $S$ during both training and inference, with the number of tokens predicted at each AR step fixed to $|\kappa_s| = \frac{L}{S}$. Specifically, we modify the input sequence to include clean image tokens and use generalized causal attention masks within the attention modules. Figure~\ref{fig:arch} illustrates the architectural differences between DiT and CausalFusion.
We train several CausalFusion models with different numbers of AR steps, i.e., $S$ = 1, 2, 4, 8, and 256. Here, $S$ = 1 indicates the In-context DiT, while $S$ = 256, equivalent to $L$, represents the pure AR training mode. To evaluate these models, we first use the same number of AR steps as in training, and further study their generalization performance to other number of AR steps. 

As shown in Table~\ref{tab:ng}, CausalFusion trained with fixed AR steps cannot be robustly transferred to other inference settings. E.g., all models yield substantially worse performance when their inference settings are not aligned with training. By comparing the best evaluation result of each training setting, we observe that increasing the number of AR steps leads to a huge decline in performance. Specifically, the 8-step CausalFusion yields an FID of 20.01, clearly lagging behind the 13.78 FID achieved by In-context DiT. However, from the loss curves in Figure~\ref{fig:weighting_and_decay}(a), an opposite trend is observed, where models trained with more AR steps consistently exhibit lower loss values than those with fewer AR steps. This suggests that the learning tasks become over-simplified as the number of AR steps increases.

\vspace{-5pt}
\paragraph{CausalFusion with random number of AR steps.} 
Additionally, we train a CausalFusion model where the number of AR steps $S$ are uniformly sampled from 1 to $L$, with $|\kappa_s|$ also randomly set at each AR step. We evaluate this model across various inference settings, same as above. As shown in Table~\ref{tab:ng}, this training setting performs relatively consistent under different inference AR steps compared to those trained with fixed AR steps, demonstrating its greater flexibility during training and versatility during inference. However, this setting still leads to inferior results compared to the In-context DiT baseline, e.g., an FID of 21.31 when evaluated with a single AR step ($S$ = 1) as diffusion mode. 
Figure~\ref{fig:weighting_and_decay}(b) offers further insight into this behavior, showing that uniform AR step sampling during training leads to a highly imbalanced $|\kappa_s|$ distribution. As a result, the training signal is dominated by AR steps with very few tokens—over 95\% of AR steps have $|\kappa_s| \leq 16$. These steps are uniformly distributed along the AR axis, causing the model to overly rely on visible context and thus diminishing the complexity of the training task.

Lastly, given the CausalFusion model trained with random AR steps, we compare the loss values produced by different AR steps on the validation set. As shown in Figure~\ref{fig:weighting_and_decay}(c), later AR steps yield much lower loss values than earlier steps, suggesting a clear trend of vanished training signals along the AR axis. 

\section{Shaping task difficulties in CausalFusion} \label{sec:difficulty}
Based on the above observations, we aim to adjust the difficulties of the generative tasks in CausalFusion to balance training signal impact and ensure thorough exploration of the factorization space. By default, we use random AR steps during training due to its effectiveness in generalizing across various inference settings. Building on this setup, we identify several straightforward approaches that effectively shape task difficulties within CausalFusion, leading to significant performance improvements. We categorize the discussion into two parts: design choices for AR step sampling and loss weighting along the AR axis. 

\vspace{-5pt}
\paragraph{Random AR steps with decayed sampling.} 
Instead of uniformly sampling the number of AR steps $S$ from $[1, L]$, we propose to exponentially decrease the sampling probability as $S$ increases, which alleviates the problem of imbalanced $|\kappa|_s$ distribution, as shown in Figure~\ref{fig:weighting_and_decay}(b).
As a result, large $|\kappa_s|$ appears more frequently in the training sequence, and more tokens are predicted base on less visible context. We introduce a hyper-parameter $\gamma$ $\leq$ 1 to control the exponential decay ratio where $\gamma$ = 1 denotes the naive CausalFusion model trained with uniformly sampled AR steps, while $\gamma$ = 0 represents our In-context DiT baseline.
From Table~\ref{tab:arweighting}(a), by decreasing $\gamma$ from 1.0 to 0.95, we obtain remarkable performance improvements across all inference settings, with gains of nearly 7 and 5 points when $S_\text{eval}$ = 1 and 2, respectively. Furthermore, when $\gamma$ = 0.9, CausalFusion surpasses the strong In-context DiT using the pure diffusion inference mode, and performance in other inference settings is further enhanced. While smaller values of $\gamma$, such as 0.8, yield even better performance with one AR step evaluation, we set the defualt value to 0.9 as it provides a balanced improvement across all inference settings.

\begin{table*}[t]
\footnotesize
\begin{tabular}{c|c|cccc|cccc|cccc}
& & \multicolumn{4}{c}{256$\times$256, w/o CFG} & \multicolumn{4}{c}{256$\times$256, w/ CFG} & \multicolumn{4}{c}{512$\times$512, w/ CFG}\\
& Params & FID$\downarrow$ & IS$\uparrow$ & Pre.$\uparrow$ & Rec.$\uparrow$ & FID$\downarrow$ & IS$\uparrow$ & Pre.$\uparrow$ & Rec.$\uparrow$ & FID$\downarrow$ & IS$\uparrow$ & Pre.$\uparrow$ & Rec.$\uparrow$ \\
\shline
\textcolor{lightgray}{GIVT~\cite{tschannen2025givt}} & \textcolor{lightgray}{304M} & \textcolor{lightgray}{5.67} & \textcolor{lightgray}{-} & \textcolor{lightgray}{0.75} & \textcolor{lightgray}{0.59} & \textcolor{lightgray}{3.35} & \textcolor{lightgray}{-} & \textcolor{lightgray}{0.84} & \textcolor{lightgray}{0.53} & \textcolor{lightgray}{2.92} & \textcolor{lightgray}{-} & \textcolor{lightgray}{0.84} & \textcolor{lightgray}{0.55} \\
\textcolor{lightgray}{MAR-B~\cite{mar}} & \textcolor{lightgray}{208M} & \textcolor{lightgray}{3.48} & \textcolor{lightgray}{192.4} & \textcolor{lightgray}{0.78} & \textcolor{lightgray}{0.58} & \textcolor{lightgray}{2.31} & \textcolor{lightgray}{281.7} & \textcolor{lightgray}{0.82} & \textcolor{lightgray}{0.57} & \textcolor{lightgray}{-} & \textcolor{lightgray}{-} & \textcolor{lightgray}{-} & \textcolor{lightgray}{-} \\
LDM-4~\cite{rombach2022high} & 400M & 10.56 & 103.5 & 0.71 & 0.62 & 3.6 & 247.7 & \textbf{0.87} & 0.48 & - & - & - & - \\
CausalFusion-L & 368M & \textbf{5.12} & \textbf{166.1} & \textbf{0.73} & \textbf{0.66} & \textbf{1.94} & \textbf{264.4} & 0.82 & \textbf{0.59} & - & - & - & - \\
\hline
\textcolor{lightgray}{MAR-L~\cite{mar}} & \textcolor{lightgray}{479M} & \textcolor{lightgray}{2.6} & \textcolor{lightgray}{221.4} & \textcolor{lightgray}{0.79} & \textcolor{lightgray}{0.60} & \textcolor{lightgray}{1.78} & \textcolor{lightgray}{296.0} & \textcolor{lightgray}{0.81} & \textcolor{lightgray}{0.60} & \textcolor{lightgray}{1.73} & \textcolor{lightgray}{279.9} & \textcolor{lightgray}{-} & \textcolor{lightgray}{-} \\
ADM~\cite{adm} & 554M & 10.94 & - & 0.69 & 0.63 & 4.59 & 186.7 & 0.82 & 0.52 & 3.85 & 221.7 & {0.84} & 0.53 \\
DiT-XL~\cite{dit} & 675M & 9.62 & 121.5 & 0.67 & 0.67 & 2.27 & 278.2 & \textbf{0.83} & 0.57 & 3.04 & 240.8 & {0.84} & 0.54 \\
SiT-XL~\cite{sit} & 675M & 8.3 & - & - & - & 2.06 & 270.3 & 0.82 & 0.59 & 2.62 & 252.2 & \textbf{0.84} & 0.57 \\
ViT-XL~\cite{minsnr} & 451M & 8.10 & - & - & -  & 2.06 & - & - & - & - & - & - & - \\
U-ViT-H/2~\cite{bao2023all} & 501M & 6.58 & - & - & - & 2.29 & 263.9 & 0.82 & 0.57 & 4.05 & - & - & -\\
MaskDiT~\cite{zheng2023fast} & 675M & 5.69 & 178.0 & 0.74 & 0.60 & 2.28 & 276.6 & 0.80 & 0.61 & 2.50 & 256.3 & 0.83 & 0.56 \\
RDM~\cite{teng2023relay} & 553M & 5.27 & 153.4 & 0.75 & 0.62 & 1.99 & 260.4 & 0.81 & 0.58 & - & - & - & - \\
CausalFusion-XL & 676M & \textbf{3.61} & \textbf{180.9} & \textbf{0.75} & \textbf{0.66} & \textbf{1.77} & \textbf{282.3} & 0.82 & \textbf{0.61} & \textbf{1.98} & \textbf{283.2} & 0.83 & \textbf{0.58} \\
\end{tabular}
\vspace{-8pt}
\caption{
\textbf{System performance comparison} on ImageNet class-conditioned generation. Numbers marked with \textcolor{lightgray}{gray} blocks use temperature sampling during inference.
\vspace{-8pt}
}
\label{tab:system}
\end{table*}

\vspace{-5pt}
\paragraph{Loss weighting along AR axis.} We modify the weighting term $w(\cdot)$ in Eqn.~(\ref{eq:diffusion_loss}) to further consider the AR step $s$. In practice, we simply set $w(s, t)$ to a pre-defined value $\lambda \geq 1$ at $s=1$, and linearly decay it to 1 at $s=S$, and keep using constant weight for different $t$. In this way, the model is trained to focus more on the hard generative tasks at early AR steps or larger noise levels.
We analysis the impact of $\lambda$ in Table~\ref{tab:arweighting}(c). From the table, setting $\lambda$ to a proper value improves the performance. Intuitively, as the model generates closer to the end of the AR axis, the task becomes easier due to high \textbf{locality}\cite{wang2018non} in the visible context, causing some generative tasks to degrade into local feature interpolation. 
In contrast, predictions made during earlier AR steps facilitate the learning of non-local dependencies within the visual context, which is beneficial to generative modeling.

\vspace{-5pt}
\paragraph{Difficulty vs. locality.} The hypothesis of {locality} aligns with our observations in Table~\ref{tab:arweighting}(b), where using random sequence order in CausalFusion during training significantly outperforms manually assigned orders. Specifically, using fixed (block-wise) raster order leads the model to rely excessively on local tokens, which makes the training task easier. In contrast, CausalFusion is trained with a random order by default, following the principle of \textit{minimal inductive bias}. This design encourages the model to develop robust generative modeling abilities rather than relying on fixed ordering priors, while allowing flexible inference orders.

\section{Performance comparison}\label{sec:sota_exp}

\begin{table*}[t]
\footnotesize
\begin{tabular}{c|ccccccc}
& Type & Tokenizer & Params & Training Epoch & Sampler (Steps) & Sampling tricks & FID$\downarrow$  \\
\shline
Open-MAGVIT2-L~\cite{luo2024open} & AR & MAGVIT2 & 800M & 300 & AR(256) & N/A  & 2.51 \\
Open-MAGVIT2-XL~\cite{luo2024open} & AR & MAGVIT2 & 1.5B & 300 & AR(256) & N/A  & 2.33 \\
LlamaGen-3B~\cite{sun2024autoregressive} & AR & custom & 3.1B & - & AR(256) & N/A & 2.18 \\
VAR-d24~\cite{tian2024visual} & VAR & custom & 1B & 350 & VAR & N/A  & 2.09\\
VAR-d30~\cite{tian2024visual} & VAR & custom & 2B & 350 & VAR & reject sampling & 1.73 \\
\hline
Simple-diffusion~\cite{hoogeboom2023simple} & Diffusion & N/A & 2B & 800 & DDPM & N/A  & 2.44 \\ 
FiTv2-3B~\cite{wang2024fitv2} & Diffusion & SD & 3B & 256 & DDPM(250) & N/A & 2.15 \\
VDM++~\cite{kingma2024understanding} & Diffusion & N/A & 2B & - & EDM & - & 2.12 \\
Large-DiT-7B~\cite{gao2024lumina} & Diffusion & SD & 3B & 435 & DDPM(250) & N/A & 2.10 \\
Flag-DiT-3B~\cite{gao2024lumina} & Diffusion & SD & 3B & 256 & adaptive Dopri-5 & N/A & 1.96 \\
DiT-MoE-XL/2-8E2A~\cite{fei2024scaling} & Diffusion & SD & 16B & $\approx$1000 & DDPM(250) & N/A & 1.72 \\
DiMR-G/2R~\cite{liu2024alleviating} & Diffusion & SD & 1.1B & 800 & DPM-solver(250) & N/A & 1.63 \\
\hline
DART-XL~\cite{dart} & AR+Diffusion & LDM & 812M & - & AR(256)+FM(100) & $\tau$ sampling & 3.98 \\
MonoFormer~\cite{monoformer} & AR+Diffusion & SD & 1.1B & - & DDPM(250) & N/A & 2.57 \\
BiGR-XL-d24~\cite{bigr} & AR+Diffusion & custom & 799M & 400 & AR(256)+DDPM(100) & $\tau$ sampling  & 2.49 \\
BiGR-XXL-d32~\cite{bigr} & AR+Diffusion & custom & 1.5B & 400 & AR(256)+DDPM(100) & $\tau$ sampling  & 2.36 \\
MAR-H~\cite{mar} & AR+Diffusion & custom & 943M & 800 & AR(256)+DDPM(100) & $\tau$ sampling & 1.55 \\
\hline
CausalFusion-H & Diffusion & custom & 1B & 800 & DDPM(250) & N/A & 1.64 \\
CausalFusion-H & Diffusion & custom & 1B & 800 & DDPM(250) & CFG interval & {1.57} \\
\end{tabular}
\vspace{-5pt}
\caption{
\textbf{System performance comparison} on 256$\times$256 ImageNet generation, compared with previously reported large models.
\vspace{-10pt}
}
\label{tab:system2}
\end{table*}

\paragraph{Class-conditional image generation.}
\label{sec:system}
We evaluate our final method on the ImageNet class-conditional generation benchmark. For system-level comparisons, we use 64 tokens to encode the class condition. The impact of varying the number of class tokens is analyzed in Appendix~\ref{appendix:secB}. We train three sizes of CausalFusion models: CausalFusion-L (368M), CausalFusion-XL (676M), and CausalFusion-H (1B). All models are trained for 800 epochs with a batch size of 2048. By default, we use a single AR step inference with 250 DDPM steps as in DiT~\cite{dit}, and report FID-50k for benchmarking against existing models. Detailed hyperparameters are provided in Appendix~\ref{appendix:secC}. As shown in Table~\ref{tab:system}, on 256$\times$256 image generation, CausalFusion-L achieves an FID-50k of 5.12 without classifier-free guidance~\cite{ho2022classifier} (CFG), outperforming DiT-XL/2 by 4.5 points with 50\% fewer parameters. CausalFusion-XL further improves this result to 3.61, and when using CFG, achieves a state-of-the-art result of 1.77, significantly outperforming strong baselines like DiT and SiT. Additionally, CausalFusion-XL demonstrates effectiveness in high-resolution generation, achieving an FID of 1.98 on 512$\times$512 images with CFG.

We also provide a system-level comparison with existing methods in Table~\ref{tab:system2}. CausalFusion-H achieves an FID of 1.64 using the standard 250-step DDPM sampler, outperforming previous diffusion models with larger model sizes, such as FiTv2-3B~\cite{wang2024fitv2} and Large-DiT-7B~\cite{gao2024lumina}, and achieving comparable results to DiMR-G/2R (1.64 vs. 1.63) despite DiMR-G/2R’s use of a stronger sampler (DPM-solver~\cite{dpm}). By applying the CFG interval~\cite{kynkaanniemi2024applying} approach, CausalFusion-H further improves to an FID of 1.57, positioning it among the top-performing models on the ImageNet 256$\times$256 benchmark.

\begin{figure}[h!]
    \centering
    \vspace{0pt}
    \begin{subfigure}{1\textwidth}
        \centering
        \includegraphics[width=\linewidth]{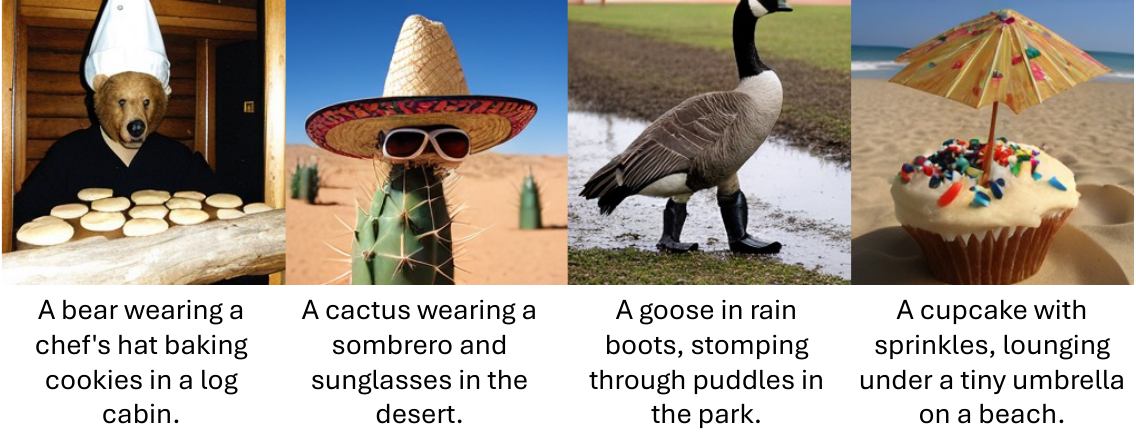}
        \caption{Samples on Text-to-Image generation.}
        \label{fig:subfig1}
    \end{subfigure}

    \begin{subfigure}{1\textwidth}
        \centering
        \includegraphics[width=\linewidth]{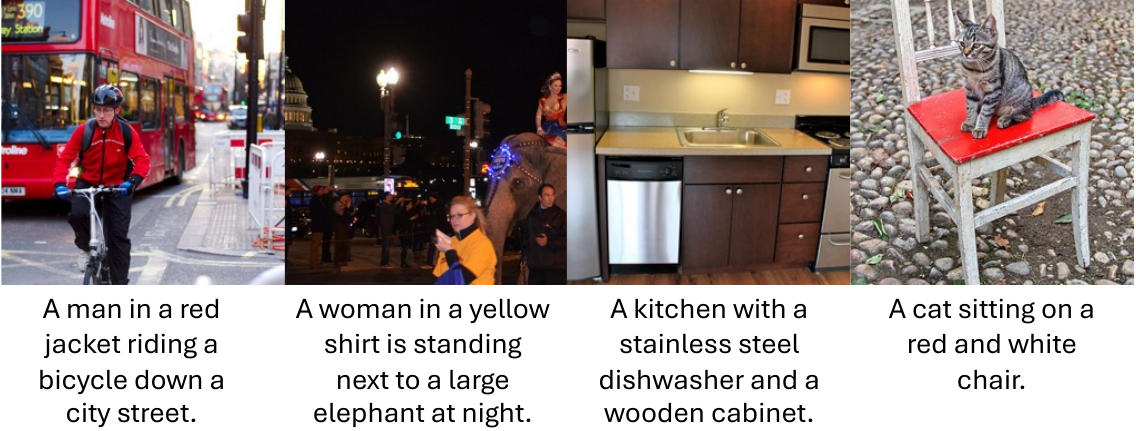}
        \caption{Samples on Image Caption generation.     \vspace{-5pt}
        }
        \label{fig:subfig2}
    \end{subfigure}
    \vspace{-5pt}
    \caption{\textbf{Multimodal generation}. Results are generated by a single CausalFusion XL model trained on ImageNet recaption data.
    \vspace{-10pt}
    }
    \label{fig:combined}
\end{figure}

\begin{table*}[h!]
\footnotesize
\centering
\subfloat[MSCOCO~\cite{lin2014microsoft} is used for FID30k and CIDEr evalution.]{
\begin{tabular}{c|cc|cc}
  & Source & Size & FID30k$\downarrow$ & CIDEr$\uparrow$ \\
\shline
Transfusion-L~\cite{transfusion} & IN1KCap & 1M & 8.1 & 34.5 \\
CausalFusion-L & IN1KCap & 1M & \textbf{7.1} & \textbf{47.9} \\
\end{tabular}
}
\hfill
\subfloat[ImageNet is used for FID10k and Acc, MSCOCO is used for CIDEr evalution.]{
\begin{tabular}{c|c|cc|cc|c}
 & Params & Data & Size & FID10k$\downarrow$ & Acc$\uparrow$ &  CIDEr$\uparrow$ \\
\shline
DiT~\cite{dit} & 458M & IN1K & 1M & 18.2 & 83.5  & 94.4 \\
CausalFusion& 368M & IN1K & 1M & 11.8 & 84.2  & 98.0 \\
{CausalFusion}$^\dagger$ & 368M  & IN1K & 1M & \textbf{9.3} & \textbf{84.7} & \textbf{103.2} \\
\end{tabular}
}
\vspace{-5pt}
\caption{
(a) \textbf{Comparison with Transfusion}~\cite{transfusion} on perception and generation benchmarks. All models are trained under the same settings using the same pretraining data. (b) \textbf{Comparison with DiT}~\cite{dit} on perception and generation benchmarks. The model marked with $\dagger$ is trained with a VAE from \cite{mar}, using a loss function to predict latent variables rather than noise.
\vspace{-10pt}
}
\label{tab:multimodal}
\end{table*}

\vspace{-5pt}
\paragraph{Zero-shot image editing.}
CausalFusion naturally supports zero-shot image editing, as it is trained to predict a random subset of image tokens based on a random subset of visible image tokens. This inherent flexibility enables the model to perform localized edits without requiring task-specific fine-tuning. As shown in Figure~\ref{fig:vis1}(b), our model can generate high-quality editing results even when only pretrained on the ImageNet class-conditional generation task, demonstrating its robustness and adaptability to editing tasks. Moreover, CausalFusion's dual-factorized design allows it to balance contextual coherence with high-fidelity updates, ensuring that edited regions blend seamlessly into the surrounding content. See Appendix~\ref{appendix:secD} for additional visualizations showcasing the model's ability to handle diverse editing scenarios.

\vspace{-5pt}
\paragraph{Vision-Language joint modeling.}
CausalFusion can integrate the language modality by applying a separate next-token prediction loss on text, similar to GPT~\cite{gpt1}, enabling it to jointly model both image and text data. In this experiment, CausalFusion was trained on two tasks simultaneously: Text-to-Image (T2I) generation and Image Captioning. During training, text precedes the image in 90\% of cases, framing it as a T2I task where only the image loss is applied. In the remaining cases, the text follows the image, and both text loss (for Image Captioning) and image loss (for classifier-free guidance~\cite{ho2022classifier} in T2I) are applied, with the text loss weighted at 0.01 relative to the image loss.

We follow the configurations and training/inference protocols from previous sections. For language tokenization, we use the LLaMA-3~\cite{llama3} tokenizer. Models are trained on a re-captioned ImageNet dataset, with each image labeled by 10 captions generated by Qwen2VL-7B~\cite{qwen2vl}. T2I and Image Captioning tasks are evaluated using zero-shot MSCOCO-30k FID and zero-shot MSCOCO CIDEr on Karpathy's test split, respectively. We compare CausalFusion with a contemporary multimodal model, Transfusion~\cite{transfusion}, which integrates language and vision modeling using standard diffusion loss for images and next-token prediction loss for text. In Transfusion, language tokens are conditioned on image embeddings with added diffusion noise. As Transfusion is not open-sourced, we implemented it based on the original paper, aligning the model architecture, VAE encoder, and language tokenizer with those used in CausalFusion.
Results with 240-epoch training are presented in Table~\ref{tab:multimodal}(a). Compared to Transfusion, CausalFusion demonstrates superior performance in both text-to-image generation and image captioning, highlighting its strong potential as a foundational model for multimodal tasks. 
In Figure \ref{fig:combined}, we show a single pretrained CausalFusion XL model performing text-to-image generation at the top and image-to-text generation (image captioning) at the bottom. Further experiment details, including data, model design, and hyperparameters, are provided in Appendix~\ref{appendix:secC}.

\vspace{-5pt}
\paragraph{Visual Representation Learning.} We further evaluate CausalFusion models from a representation learning perspective. Specifically, we leverage the CausalFusion model pretrained on the 256$\times$256 ImageNet class-conditional generation task and fine-tune it on the ImageNet classification and MSCOCO captioning tasks. For image classification, we use the average-pooled features from the last layer of CausalFusion, followed by a linear classifier. For image captioning, we add a small transformer encoder-decoder module as the language head on top of CausalFusion.
The pretrained CausalFusion model follows the default configuration described in previous sections. We compare it to the DiT-L/2 model pretrained for the same number of epochs. Detailed hyperparameters for fine-tuning are provided in Appendix~\ref{appendix:secC}. 
As shown in Table~\ref{tab:multimodal}(b), our CausalFusion model outperforms DiT on all fine-tuning tasks, indicating that CausalFusion learns superior representations compared to DiT. We hypothesize that the random grouped token diffusion mechanism in CausalFusion, which diffuses images with partially observed inputs, implicitly performs as masked representation prediction \cite{he2022masked, maskedpredict}, enhancing the model's representation learning ability.

%% file: sec/5_conclusion.tex
\vspace{-3pt}
\section{Conclusion}
We propose CausalFusion, a decoder-only transformer that unifies AR and diffusion paradigms through a dual-factorized framework across sequential tokens and diffusion noise levels. This approach achieves state-of-the-art performance on the ImageNet generation benchmark, supports arbitrary-length token generation, and enables smooth transitions between AR and diffusion modes. CausalFusion also demonstrates multimodal capabilities, including joint image generation and captioning, as well as zero-shot image manipulations. Our framework offers a new perspective on unified learning of diffusion and AR models.

%% file: sec/appendix.tex
\appendix

\section{Generalized Causal Attention}\label{appendix:secA}
\begin{figure}[h]\centering
\includegraphics[width=0.7\linewidth]{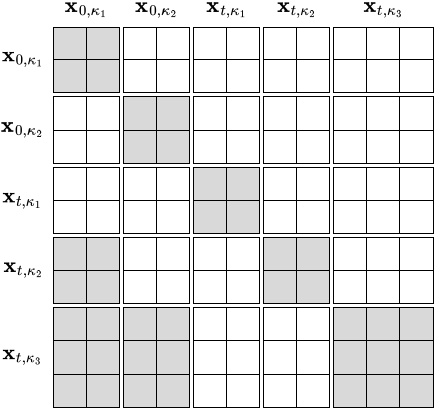}
\vspace{-5pt}
\caption{\textbf{Generalized causal mask.} In this case, the input sequence is organized to have 3 AR-steps $\kappa_1$, $\kappa_2$, and $\kappa_3$, containing 2, 2, and 3 tokens, respectively. $\mathbf{x}_{0, \kappa_1}$ and $\mathbf{x}_{0, \kappa_2}$ are the clean tokens at the first two AR steps, while $\mathbf{x}_{t, \kappa_1}$, $\mathbf{x}_{t, \kappa_2}$, and $\mathbf{x}_{t, \kappa_3}$ are noised tokens. White and gray blocks denote the masked and unmasked attention, respectively. Note that, each $\mathbf{x}_{t, \kappa_s}$ only attends to itself and the clean tokens from previous AR steps $\mathbf{x}_{0, \kappa_{1:s-1}}$.}
\label{fig:generalized_causal_mask}
\end{figure}

We design the Generalized Causal Attention for CausalFusion models. The core idea is to maintain causal dependencies across all AR steps while ensuring that each AR step relies only on \textit{clean} image tokens from preceding AR steps. This design allows CausalFusion to generate images using a \textit{next-token(s) diffusion} paradigm.
We show an example of the generalized causal attention mask in Figure~\ref{fig:generalized_causal_mask}, and the PyTorch-style pseudo code for obtaining generalized causal mask in Algorithm~\ref{algo:gcm}.

\vspace{-10pt}

\section{More Analyses}\label{appendix:secB}
\paragraph{Diffusion time steps sampling.}
Following DiT~\cite{dit}, we randomly sample the diffusion time step $t$ during training. By default, the same $t$ is used across all AR steps when training CausalFusion models. Here, we explore the impact of using different $t$ values for different AR steps during training. The training and evaluation settings remain consistent with Section~\ref{sec:init_exps}. As shown in Table~\ref{tab:timestep}, using either shared or random $t$ values results in similar performance, indicating that CausalFusion is robust to this variation.
Additionally, we evaluate a setting where multiple diffusion time steps are sampled for each AR step. Specifically, we experiment with sampling 4 and 8 different time steps, reducing the batch size by factors of 4$\times$ and 8$\times$, respectively, to ensure the total number of tokens used for loss calculation remains constant. As shown in the table, using multiple diffusion time steps per AR step achieves comparable performance to the default setting, further demonstrating the robustness of CausalFusion to this factor.
Notably, in this approach, the clean image tokens $\mathbf{x}_{0, \kappa{s}}$ at each AR step need to be computed only once and can be shared across multiple $\mathbf{x}_{t, \kappa_s}$ with different $t$ values. Consequently, the additional computational cost introduced by clean image tokens during training is minimized, amounting to only $\sim$10\%.
\vspace{-5pt}

\begin{algorithm}[t]
\caption{
Generalized causal mask}
\label{algo:gcm}
\definecolor{codeblue}{rgb}{0.25,0.5,0.5}
\definecolor{codekw}{rgb}{0.85, 0.18, 0.50}
\begin{lstlisting}[language=python]
def get_attn_mask(ctx_len, x_len, step):
    # tx_len: the length of clean tokens
    # x_len: the length of noisy tokens
    # step: number of AR steps

    # sample random tokens per AR step
    sumk = random.sample(range(1, x_len), step - 1)
    sumk = [0] + sorted(sumk) + [x_len]

    # build `causal` masks
    seq_len = ctx_len + x_len
    attn_mask = torch.ones(size=(seq_len, seq_len))
    m1 = torch.ones(size=(ctx_len, ctx_len))
    m2 = torch.ones(size=(x_len, ctx_len))
    m3 = torch.ones(size=(x_len, x_len))
    for i in range(len(sumk) - 2):
        m1[sumk[i]:sumk[i+1], 0:sumk[i+1]] = 0
        m2[sumk[i+1]:sumk[i+2], 0:sumk[i+1]] = 0
    for i in range(len(sumk) - 1):
        m3[sumk[i]:sumk[i+1], sumk[i]:sumk[i+1]] = 0

    attn_mask[:ctx_len, :ctx_len] = m1
    attn_mask[ctx_len:, :ctx_len] = m2
    attn_mask[ctx_len:, ctx_len:] = m3
    return attn_mask # 1 for mask, 0 for unmask
\end{lstlisting}
\end{algorithm}

\begin{table}[h]
\footnotesize
\centering
\begin{tabular}{c|c}
& FID10k\\ 
\shline
\underline{shared $t$ for different AR steps} & 12.13 \\
random $t$ for different AR steps & 12.27 \\
4$\times$ $t$ for each AR step & 12.19 \\
8$\times$ $t$ for each AR step & 12.23 \\
\end{tabular}
\vspace{-5pt}
\caption{\textbf{Diffusion time steps sampling} strategy does not affect the performance. The default setting is \underline{underlined}.}
\label{tab:timestep}
\end{table}
\vspace{-10pt}

\begin{table}[h]
\footnotesize
\centering
\begin{tabular}{c|cc}
\#class tokens & params (M) & FID10k\\ 
\shline
\underline{4} & 308 (+3.9) &  12.13 \\
16 & 320 (+15.6) &  12.04 \\
{64} & 368 (+62.5) &  11.84 \\
1 (repeat 64$\times$) & 305 (+1.0) & 12.29  \\
4 (repeat 16$\times$) & 308 (+ 3.9) & \textbf{11.75}  
\end{tabular}
\vspace{-5pt}
\caption{\textbf{\#Class tokens} offers a trade-off between performance and number of parameters. The default setting is \underline{underlined}.}
\label{tab:cls-token}
\end{table}
\vspace{-15pt}

\paragraph{Class condition tokens.} As discussed in Sections~\ref{sec:init_exps} and \ref{sec:sota_exp}, we use 4 class condition tokens for ablation studies and 64 tokens for system-level comparisons. Here, we examine the impact of the number of class tokens in the CausalFusion framework. As shown in Table~\ref{tab:cls-token}, increasing the number of class tokens to 64 slightly improves performance (12.13 vs. 11.84 FID). However, this also adds 62.5M parameters, a significant increase (20\%) for a CausalFusion-L model with 304M parameters. To address this, we adopt a token-repeating strategy from~\cite{mar}, which achieves comparable performance (11.75 vs. 11.84 FID) without increasing the parameter count. This finding suggests that the computation allocated to class conditioning is more critical than the number of parameters dedicated to it.

\section{Implementation Details}\label{appendix:secC}
\label{appendix:impl}

\paragraph{Class-conditional image generation.} In Table~\ref{tab:impl_abla}, we provide the detailed settings of CausalFusion models for class-conditional image generation in Section~\ref{sec:init_exps} and \ref{sec:difficulty}.
\vspace{-5pt}

\begin{table}[h]
    \footnotesize
    \begin{tabular}{c|c}
        config & value \\
        \shline
        image resolution & 256$\times$256 \\
        hidden dimension & 1024 \\
        \#heads & 16 \\
        \#layers & 24 \\
        \#cls tokens & 4 \\
        patch size & 2 \\
        positional embedding & sinusoidal \\
        VAE & SD~\cite{SD-vae} \\
        VAE donwsample& 8$\times$ \\
        latent channel & 4 \\
        \hline
        optimizer & AdamW~\cite{loshchilov2017decoupled} \\
        base learning rate & 1e-4 \\
        weight decay & 0.0 \\
        optimizer momentum & $\beta_1, \beta_2{=}0.9, 0.95$ \\
        batch size & 2048 \\
        learning rate schedule & constant \\
        warmup epochs & 40 \\
        training epochs & 240 \\
        augmentation & horizontal flip, center crop \\
        \hline
        diffusion sampler & DDPM~\cite{ddpm} \\
        diffusion steps & 250 \\
        evaluation suite & ADM~\cite{adm} \\
        evaluation metric & FID-10k
    \end{tabular}
    \caption{{Ablation study} configuration.}
    \label{tab:impl_abla} 
\end{table}

\vspace{-15pt}
\paragraph{System-level comparisons.}
In Table~\ref{tab:impl_sys}, we provide the detailed settings of CausalFusion models for system-level comparisons in Section~\ref{sec:sota_exp}.
\vspace{-5pt}

\begin{table}[h]
    \footnotesize
    \begin{tabular}{c|c}
        config & value \\
        \shline
        hidden dimension & 1024 (L), 1280 (XL), 1408 (H) \\
        \#heads & 16 (L), 20 (XL), 22 (H)  \\
        \#layers & 24 (L), 32 (XL), 40 (H) \\
        \#cls tokens & 64 \\
        positional embedding & learnable \\
        VAE & mar~\cite{mar} \\
        VAE donwsample& 16$\times$ \\
        latent channel & 16 \\
        \hline
        optimizer & AdamW~\cite{loshchilov2017decoupled} \\
        base learning rate & 1e-4 \\
        weight decay & 0.0 \\
        optimizer momentum & $\beta_1, \beta_2{=}0.9, 0.95$ \\
        batch size & 2048 \\
        learning rate schedule & constant \\
        warmup epochs & 40 \\
        training epochs & 800 \\
        augmentation & horizontal flip, center crop \\
        \hline
        diffusion sampler & DDPM~\cite{ddpm} \\
        diffusion steps & 250 \\
        evaluation suite & ADM~\cite{adm} \\
        evaluation metric & FID-50k
    \end{tabular}
    \caption{{System-level comparison} configuration.}
    \label{tab:impl_sys}
\end{table}

\vspace{-15pt}
\paragraph{Multi-modal CausalFusion.} 
\label{appendix:t2i2t}
In Table \ref{tab:t2i2t_abla}, we provide the detail experiment hyperparameters for both CausalFusion and Transfusion experiments in Section~\ref{sec:sota_exp}.
The training dataset is augmented with 10 captions per image from ImageNet, generated using Qwen2VL-7B-Instruct \cite{qwen2vl} with the following prompt:

\textit{
You are an image captioner. You need to describe images in COCO style. COCO style is short. Here are some examples of COCO style descriptions: 
`A car that seems to be parked illegally behind a legally parked car'
`This is a blue and white bathroom with a wall sink and a lifesaver on the wall.'
`Meat with vegetable and fruit displayed in roasting pan.'
`Group of men playing baseball on an open makeshift field.'
}
\vspace{-5pt}

\begin{table}[h]
    \footnotesize
    \begin{tabular}{c|c}
        config & value \\
        \shline
        image resolution & 256$\times$256 \\
        hidden dimension & 1024 \\
        \#heads & 16 \\
        \#layers & 24 \\
        \#max text tokens & 35 \\
        patch size & 2 \\
        image positional embedding & sinusoidal \\
        text positional embedding & learnable \\
        VAE & SD~\cite{SD-vae} \\
        VAE donwsample& 8$\times$ \\
        latent channel & 4 \\
        \hline
        optimizer & AdamW~\cite{loshchilov2017decoupled} \\
        base learning rate & 1e-4 \\
        text loss coefficient & 0.01 \\
        weight decay & 0.0 \\
        optimizer momentum & $\beta_1, \beta_2{=}0.9, 0.95$ \\
        batch size & 2048 \\
        learning rate schedule & constant \\
        warmup epochs & 40 \\
        training epochs & 240 \\
        augmentation & horizontal flip, center crop \\
        \hline
        diffusion sampler & DDPM~\cite{ddpm} \\
        diffusion steps & 250 \\
        generation eval. metric & MSCOCO 0-shot FID-30k \\
        captioning eval. metric & MSCOCO CIDEr (Karpathy test) \\
    \end{tabular}
    \caption{Multi-modal experiment configuration for both CausalFusion and Transfusion.}
    \label{tab:t2i2t_abla} 
\end{table}
 \vspace{-10pt}

\paragraph{Fine-tuning for ImageNet classification.}
\label{appendix:in1k}
When fine-tuning our CausalFusion model for ImageNet classification, we adhere to the basic architecture of the Vision Transformer (ViT) \cite{vit}, with the exception of the class token. 
We exclude the extra timestep embedding, label embedding, and conditional position embedding. 
Layer normalization and a linear classification layer are applied to the averaged output tokens. 
Regarding hyperparameters, we follow the MAE training recipe \cite{he2022masked} as detailed in Table \ref{tab:impl_finetune}, but we use BFloat16 precision during training to enhance stability.

\begin{table}[t]
    \footnotesize
    \begin{tabular}{c|c}
        config & value \\
        \shline
        optimizer & AdamW \\
        base learning rate & 1e-3 (L) \\
        weight decay & 0.05 \\
        optimizer momentum & $\beta_1, \beta_2{=}0.9, 0.999$ \\
        layer-wise lr decay \cite{electra,beit} & 0.85 (L) \\
        batch size & 1024 \\
        learning rate schedule & cosine decay \\
        warmup epochs & 5 \\
        training epochs & 50 (L) \\
        augmentation & RandAug (9, 0.5) \cite{randaugment} \\
        label smoothing \cite{label_smooth} & 0.1 \\
        erasing \cite{erasing} & 0.25 \\
        mixup \cite{mixup} & 0.8 \\
        cutmix \cite{cutmix} & 1.0 \\
        drop path \cite{droppath} & 0.1 (L) \\
    \end{tabular}
    \caption{{ImageNet classification end-to-end fine-tuning setting.}}
    \label{tab:impl_finetune}
\end{table}

\paragraph{Fine-tuning for MSCOCO captioning.}
We follow the COCO caption fine-tuning setup of FLIP \cite{flip}, incorporating an additional caption head consisting of a 3-layer transformer encoder and a 3-layer transformer decoder (with a width of 384 and 6 attention heads). This caption head takes image features from CausalFusion or DiT as input.
We evaluate image features from the 14th, 21st, and 24th layers of CausalFusion and DiT, selecting the layer that achieves the highest performance.
The models are fine-tuned on the Karpathy training split for 20 epochs.

\begin{table}[t]
    \footnotesize
    \begin{tabular}{c|c}
        config & value \\
        \shline
        optimizer & AdamW \\
        caption head lr   & 1e-4 \\
        other parameters lr  & 1e-5 \\
        weight decay & 0.01 \\
        dropout & 0.1 \\
        optimizer momentum & $\beta_1, \beta_2{=}0.9, 0.999$ \\
        batch size & 256 \\
        learning rate schedule & cosine decay \\
        warmup epochs & 2 \\
        training epochs & 20 \\
    \end{tabular}
    \caption{{MSCOCO captioning end-to-end fine-tuning setting}}
    \label{tab:msc_cap}
\end{table}

\section{Additional Samples}\label{appendix:secD}
\label{sec:model_sample}
\begin{figure}\centering
\includegraphics[width=\linewidth]{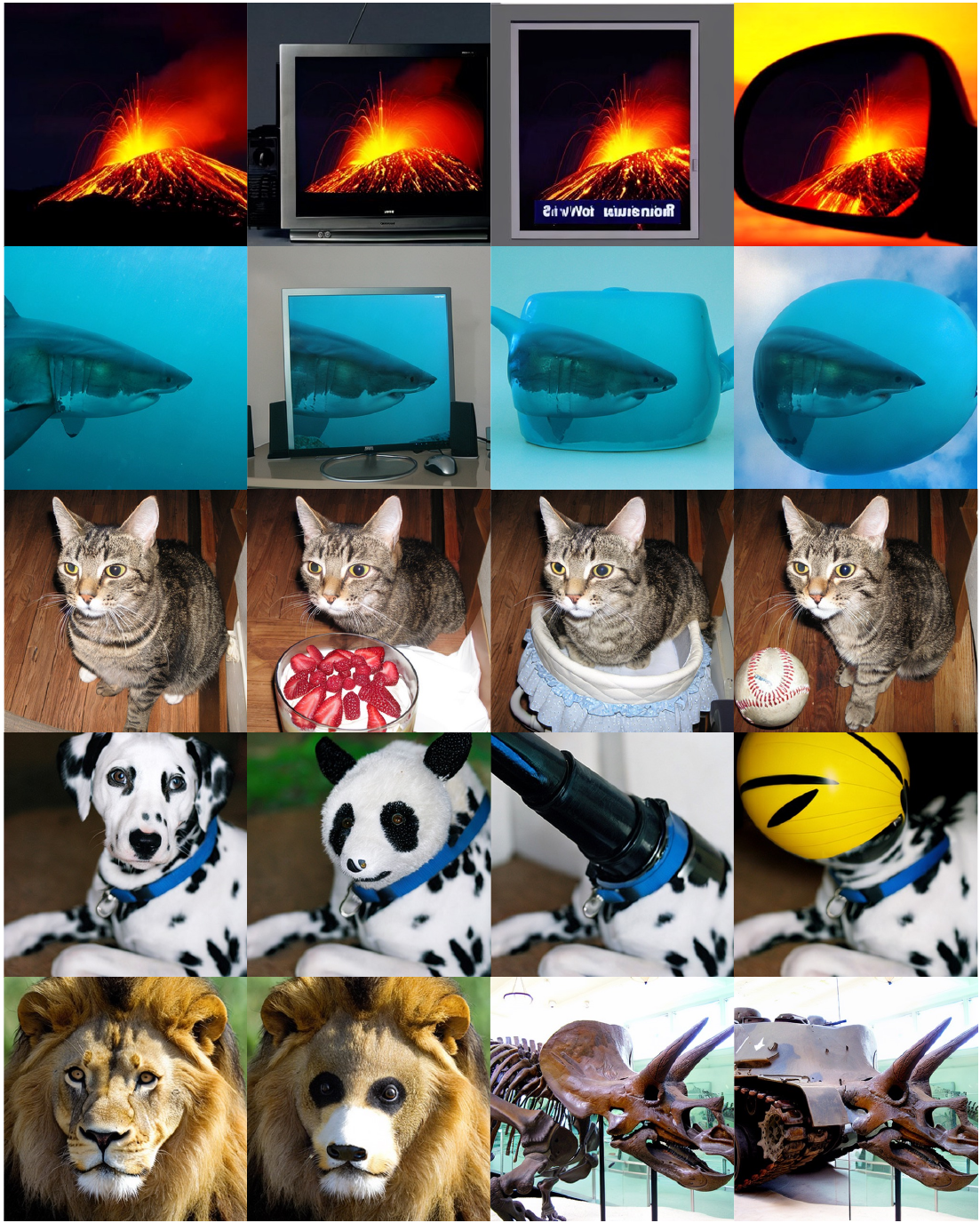}
\caption{\textbf{Zero-shot editing samples.} CausalFusion-XL, resolution 512$\times$512, 800 epoch, Classifier-free guidance scale = 3.0.}
\label{fig:edit512}
\end{figure}

\begin{figure}\centering
\includegraphics[width=\linewidth]{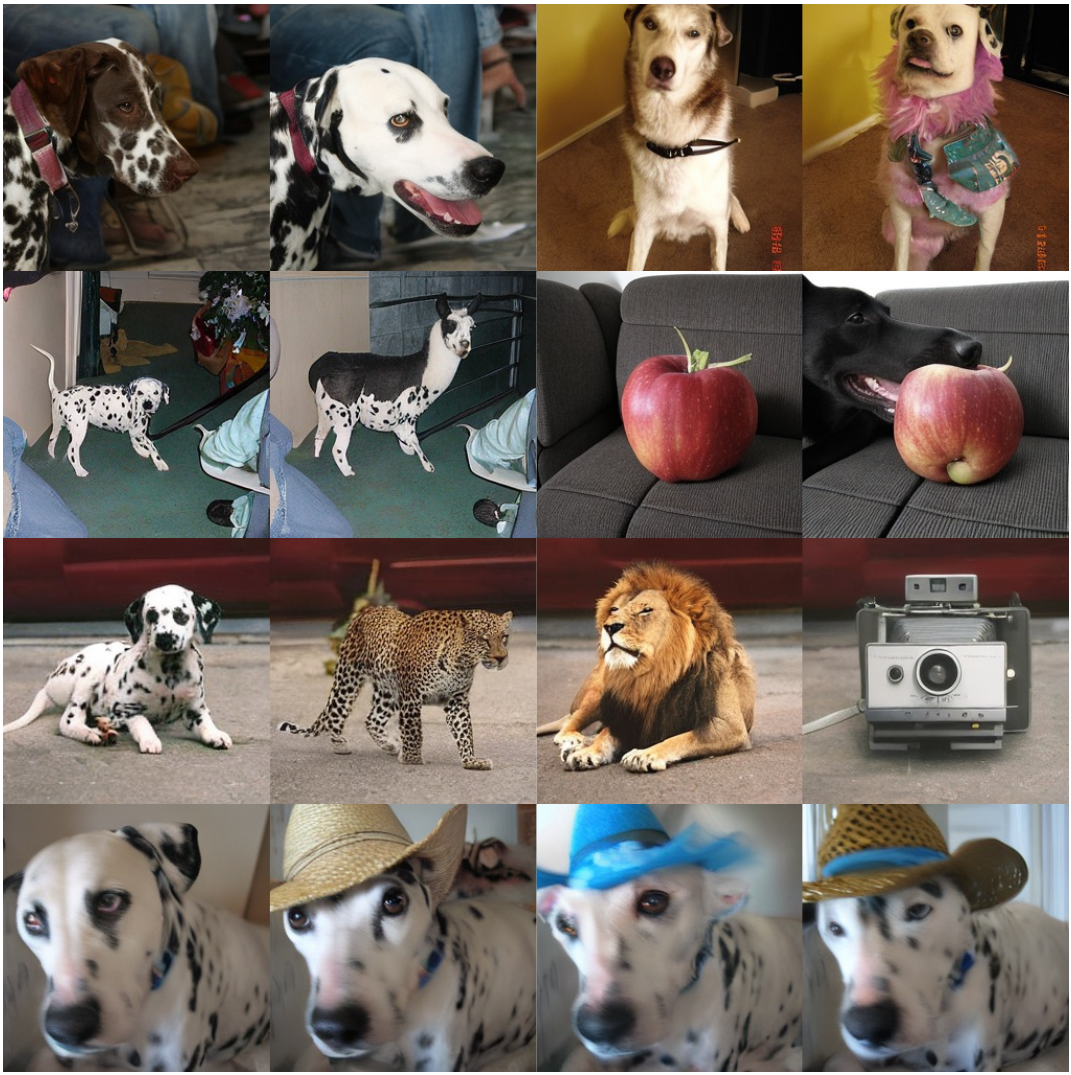}
\caption{\textbf{Zero-shot editing samples.} CausalFusion-XL, resolution 256$\times$256, 800 epoch, Classifier-free guidance scale = 1.5.}
\label{fig:edit256}
\end{figure}

We show more zero-shot editing results from our CausalDiffusion models in Figure~\ref{fig:edit512} and \ref{fig:edit256}.
The editing results are achieved by first generating the original image using the initial class label, then masking a portion of the image, and regenerating it conditioned on the unmasked region and the new class label.
For example, in the first example in Figure~\ref{fig:edit512}, an image of ``volcano'' is first generated. Then, the outer region of the image is masked out, and the new images are regenerated with new labels, such as ``televison'', ``sliding door'', and ``car mirror''.

\begin{figure}\centering
\includegraphics[width=\linewidth]{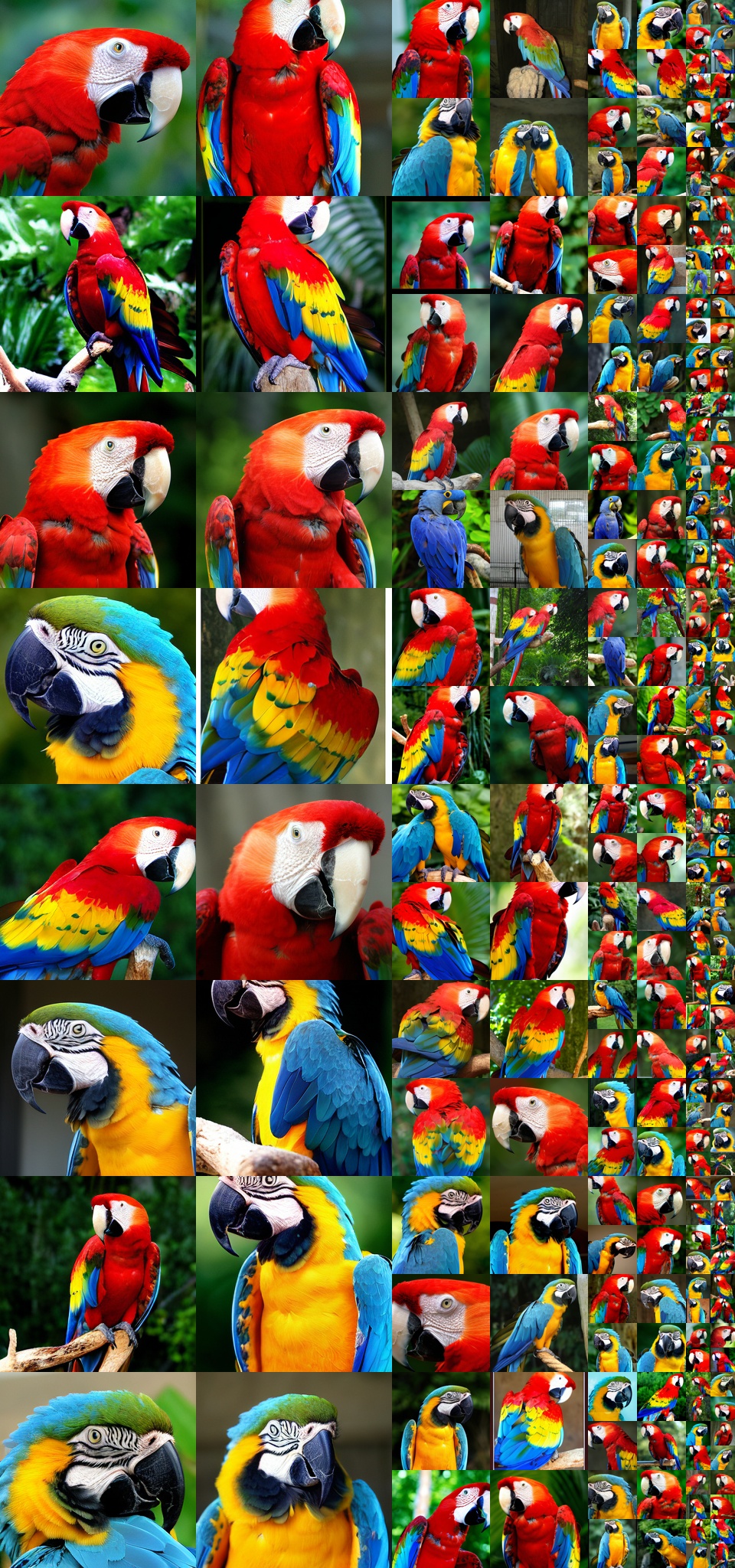}
\caption{\textbf{Uncurated $256\times256$ CausalFusion-XL samples.} \\Classifier-free guidance scale = 4.0\\Class label = ``macaw" (88)}\vspace{-2mm}
\label{fig:samples256_1}
\end{figure}

\begin{figure}\centering
\includegraphics[width=\linewidth]{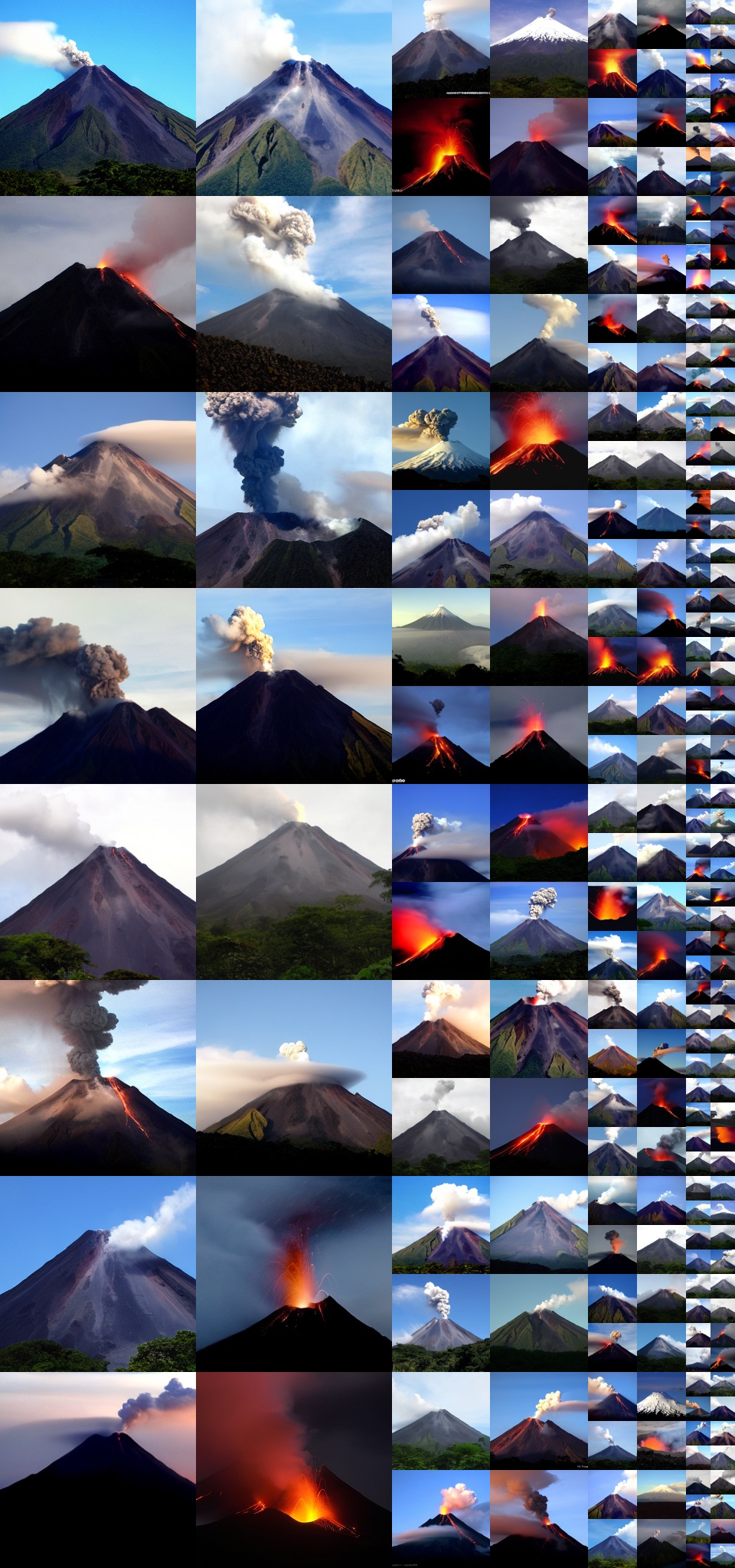}
\caption{\textbf{Uncurated $256\times256$ CausalFusion-XL samples.} \\Classifier-free guidance scale = 4.0\\Class label = ``volcano" (980)}\vspace{-2mm}
\label{fig:samples256_2}
\end{figure}

\begin{figure}\centering
\includegraphics[width=\linewidth]{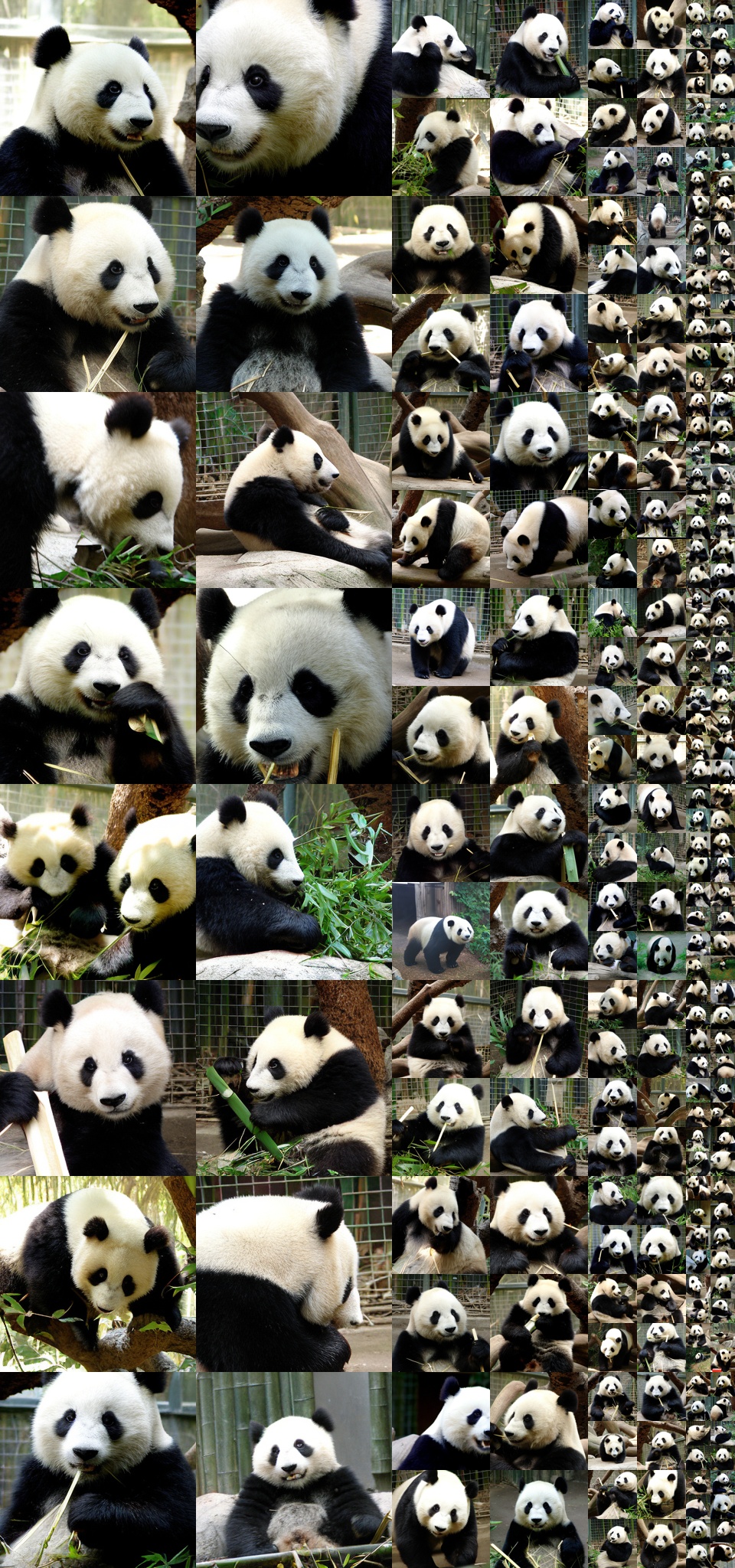}
\caption{\textbf{Uncurated $256\times256$ CausalFusion-XL samples.} \\Classifier-free guidance scale = 2.0\\Class label = ``giant panda" (388)}\vspace{-2mm}
\label{fig:samples256_3}
\end{figure}

\begin{figure}\centering
\includegraphics[width=\linewidth]{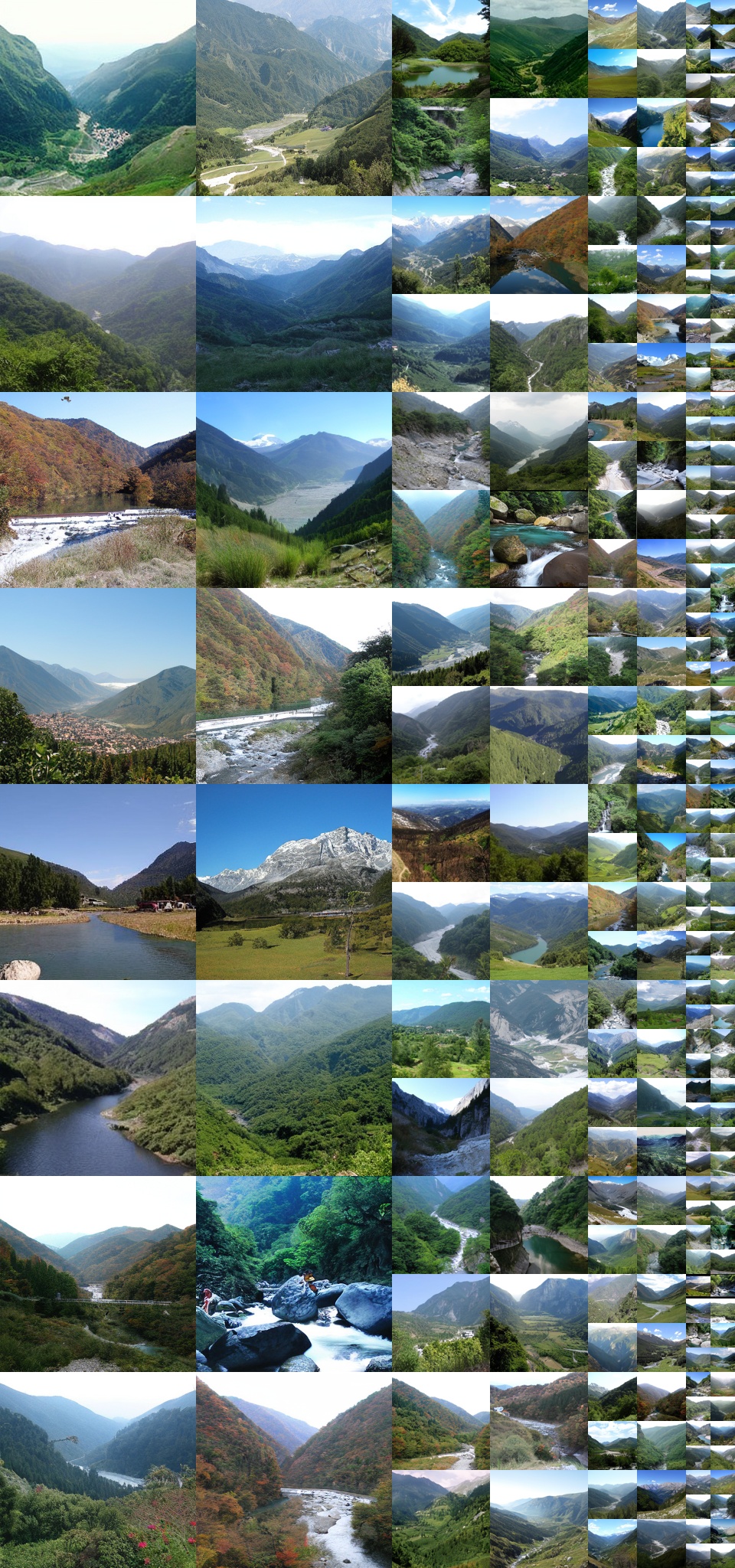}
\caption{\textbf{Uncurated $256\times256$ CausalFusion-XL samples.} \\Classifier-free guidance scale = 2.0\\Class label = ``valley" (979)}\vspace{-2mm}
\label{fig:samples256_4}
\end{figure}

\begin{figure}\centering
\includegraphics[width=\linewidth]{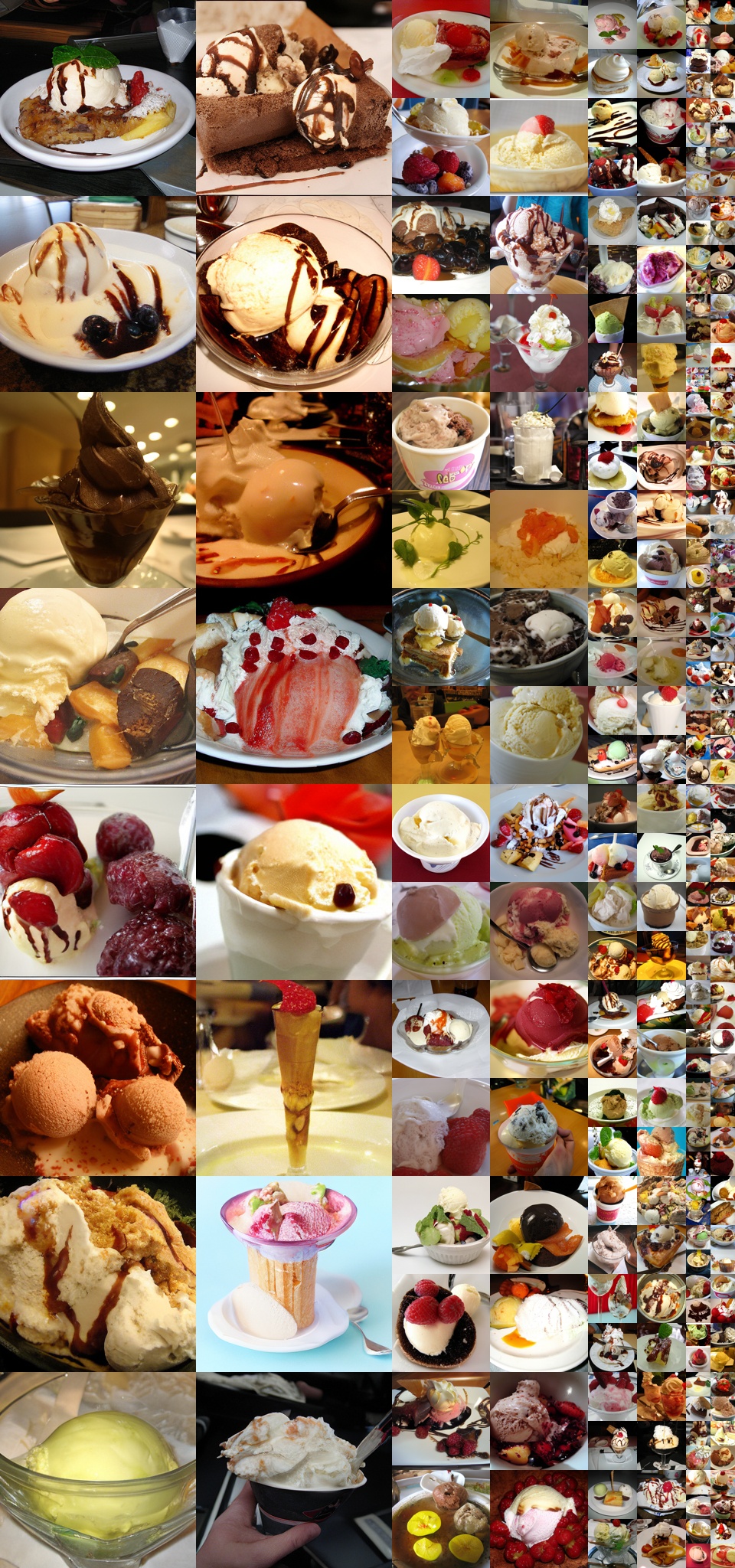}
\caption{\textbf{Uncurated $256\times256$ CausalFusion-XL samples.} \\Classifier-free guidance scale = 1.5\\Class label = ``ice cream" (928)}\vspace{-2mm}
\label{fig:samples256_5}
\end{figure}

\begin{figure}\centering
\includegraphics[width=\linewidth]{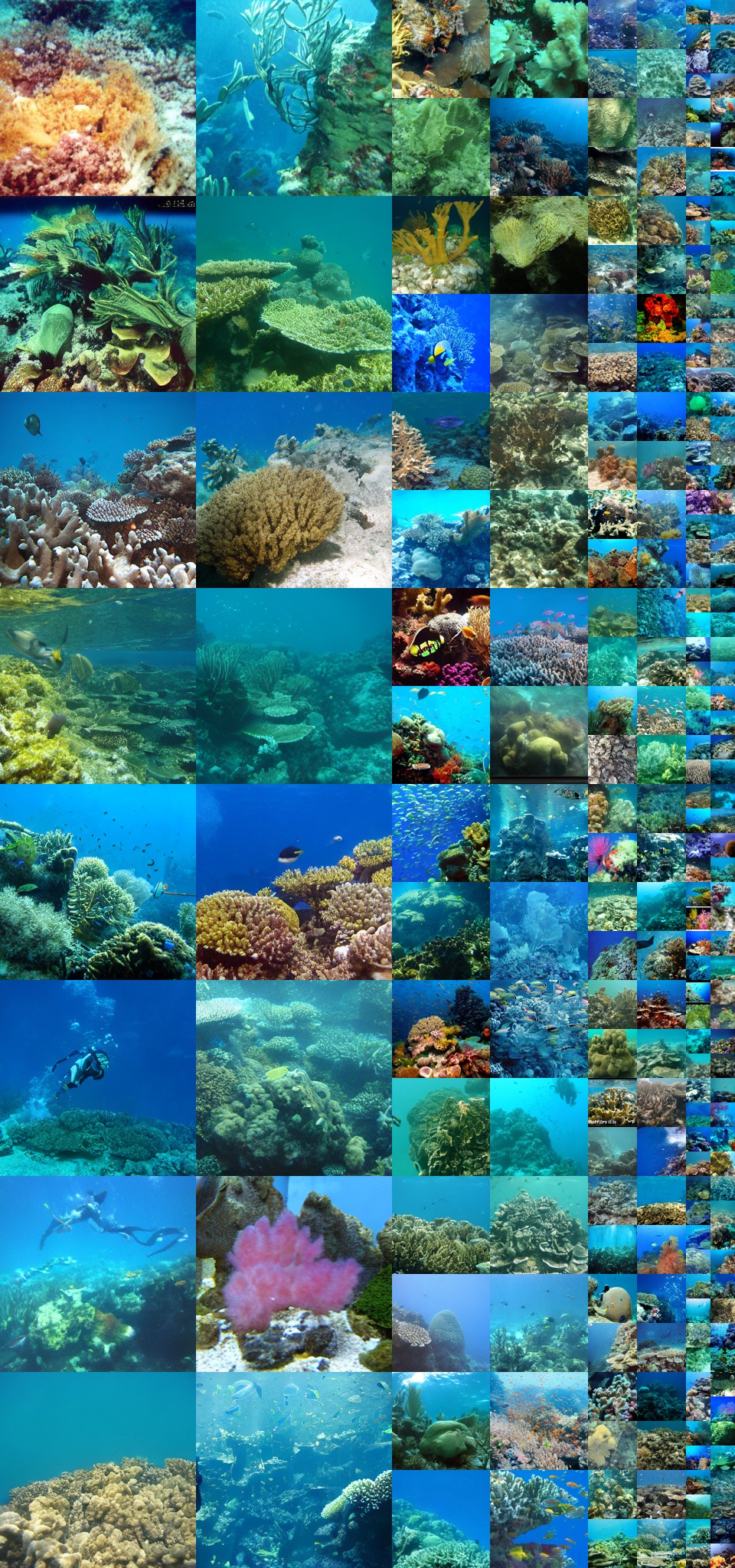}
\caption{\textbf{Uncurated $256\times256$ CausalFusion-XL samples.} \\Classifier-free guidance scale = 1.5\\Class label = ``coral reef" (973)}\vspace{-2mm}
\label{fig:samples256_6}
\end{figure}

We further show \textit{uncurated} samples from our CausalDiffusion-XL models at 512$\times$512 and 256$\times$256 resolution.
Figures \ref{fig:samples256_1} through \ref{fig:samples256_6} display samples under varying classifier-free guidance scales and class labels.

\clearpage
\begin{figure}\centering
\includegraphics[width=\linewidth]{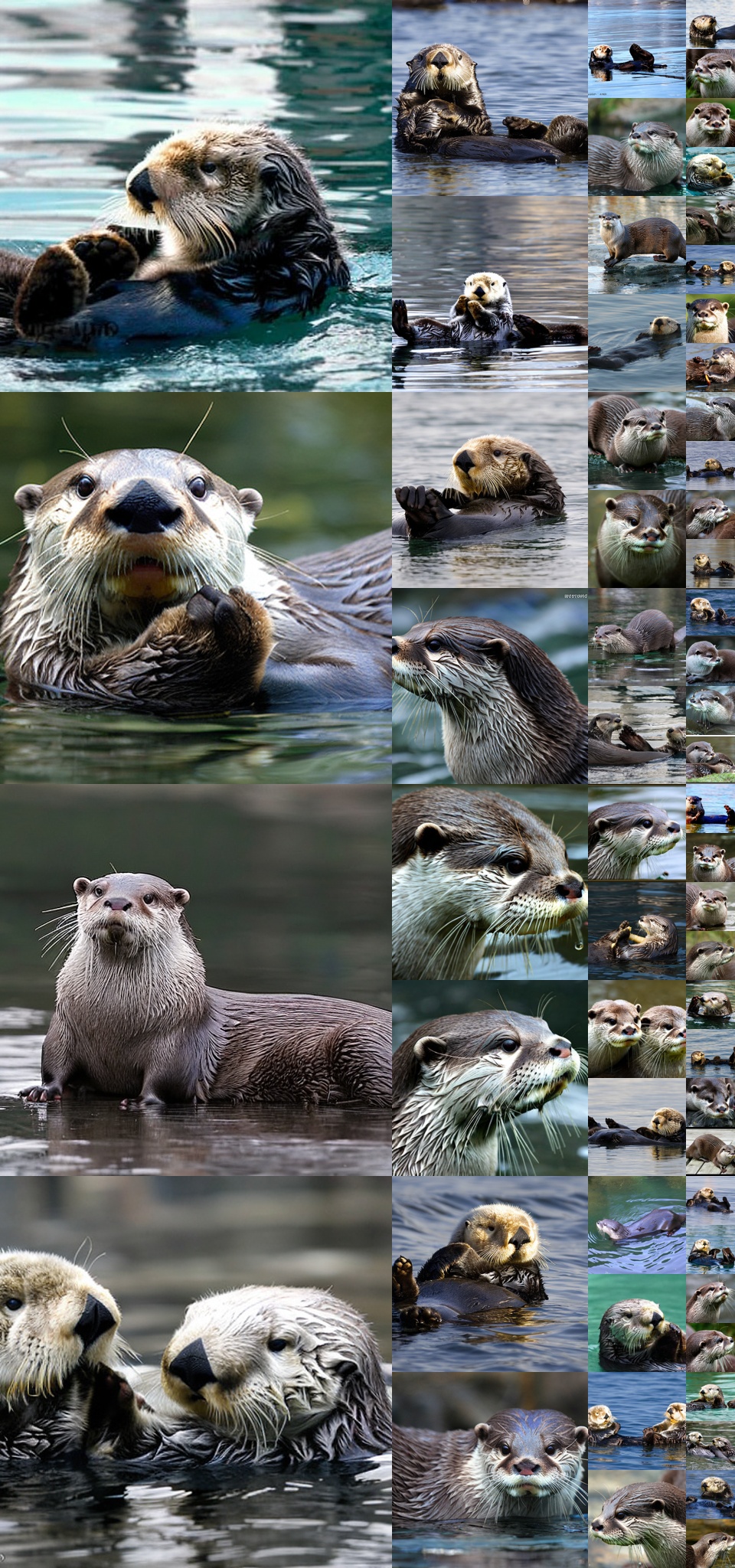}
\caption{\textbf{Uncurated $512\times512$ CausalFusion-XL samples.} \\Classifier-free guidance scale = 4.0\\Class label = ``otter" (360)}\vspace{-2mm}
\label{fig:samples512_1}
\end{figure}

\begin{figure}\centering
\includegraphics[width=\linewidth]{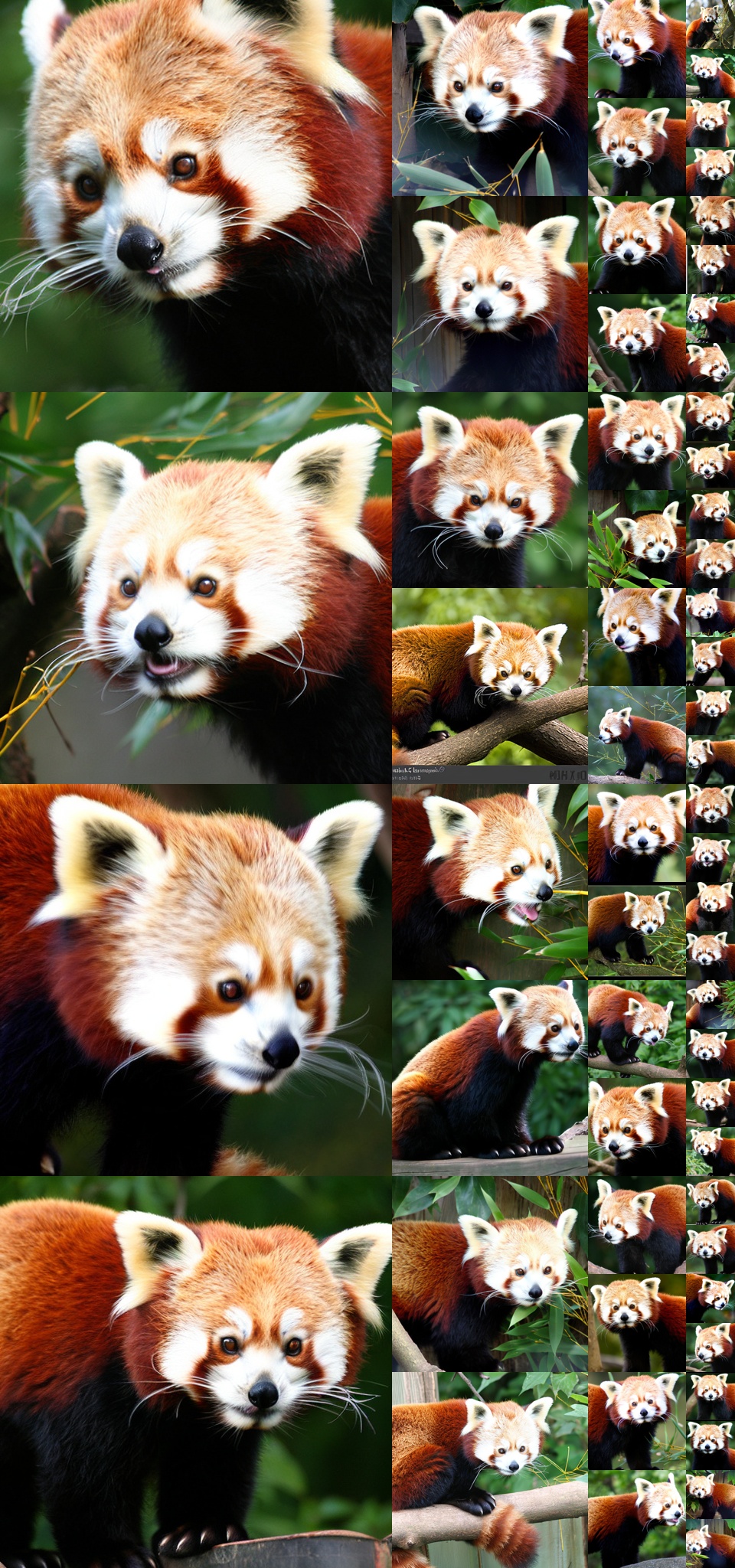}
\caption{\textbf{Uncurated $512\times512$ CausalFusion-XL samples.} \\Classifier-free guidance scale = 4.0\\Class label = ``red panda" (387)}\vspace{-2mm}
\label{fig:samples512_2}
\end{figure}

\begin{figure}\centering
\includegraphics[width=\linewidth]{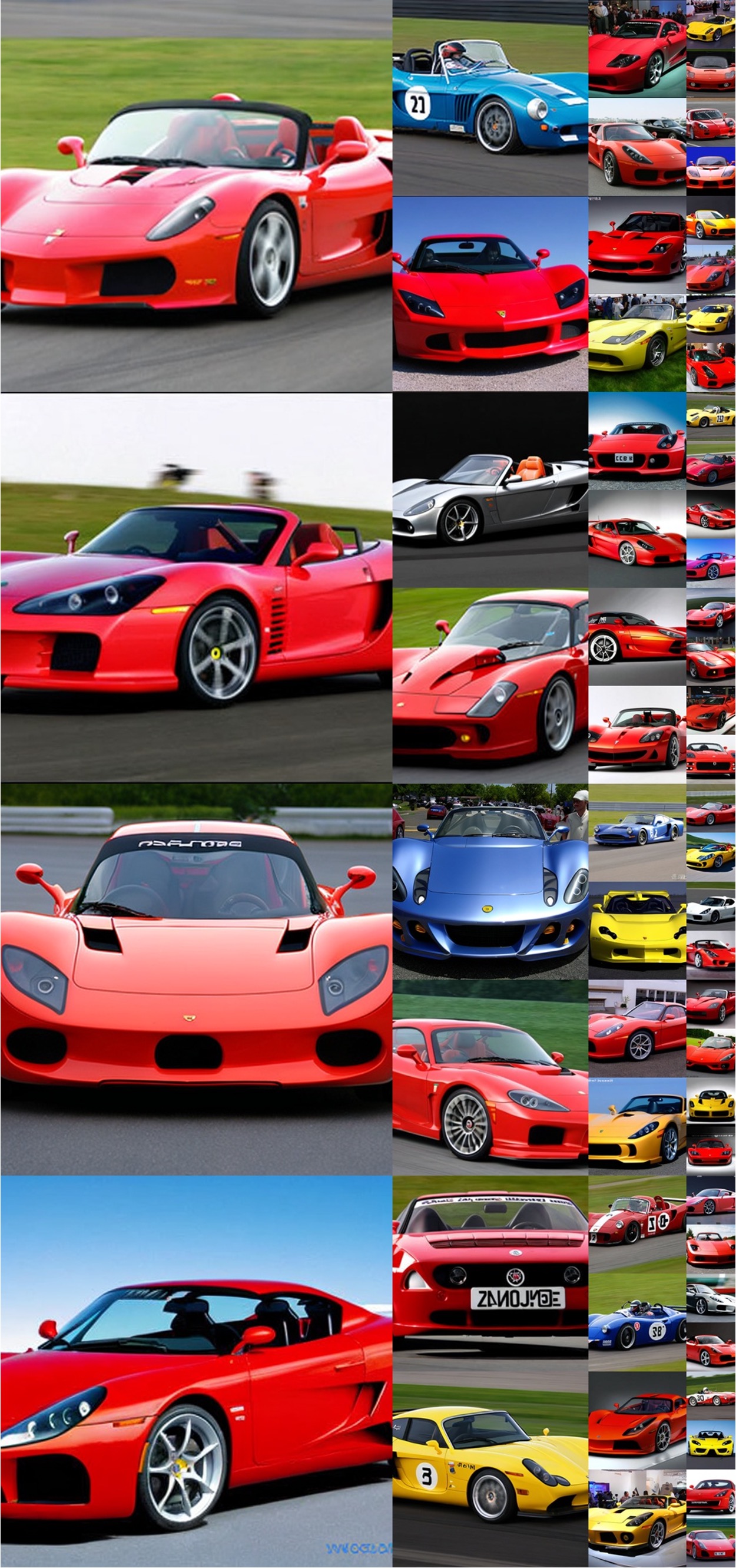}
\caption{\textbf{Uncurated $512\times512$ CausalFusion-XL samples.} \\Classifier-free guidance scale = 4.0\\Class label = ``sports car" (817)}\vspace{-2mm}
\label{fig:samples512_3}
\end{figure}

\begin{figure}\centering
\includegraphics[width=\linewidth]{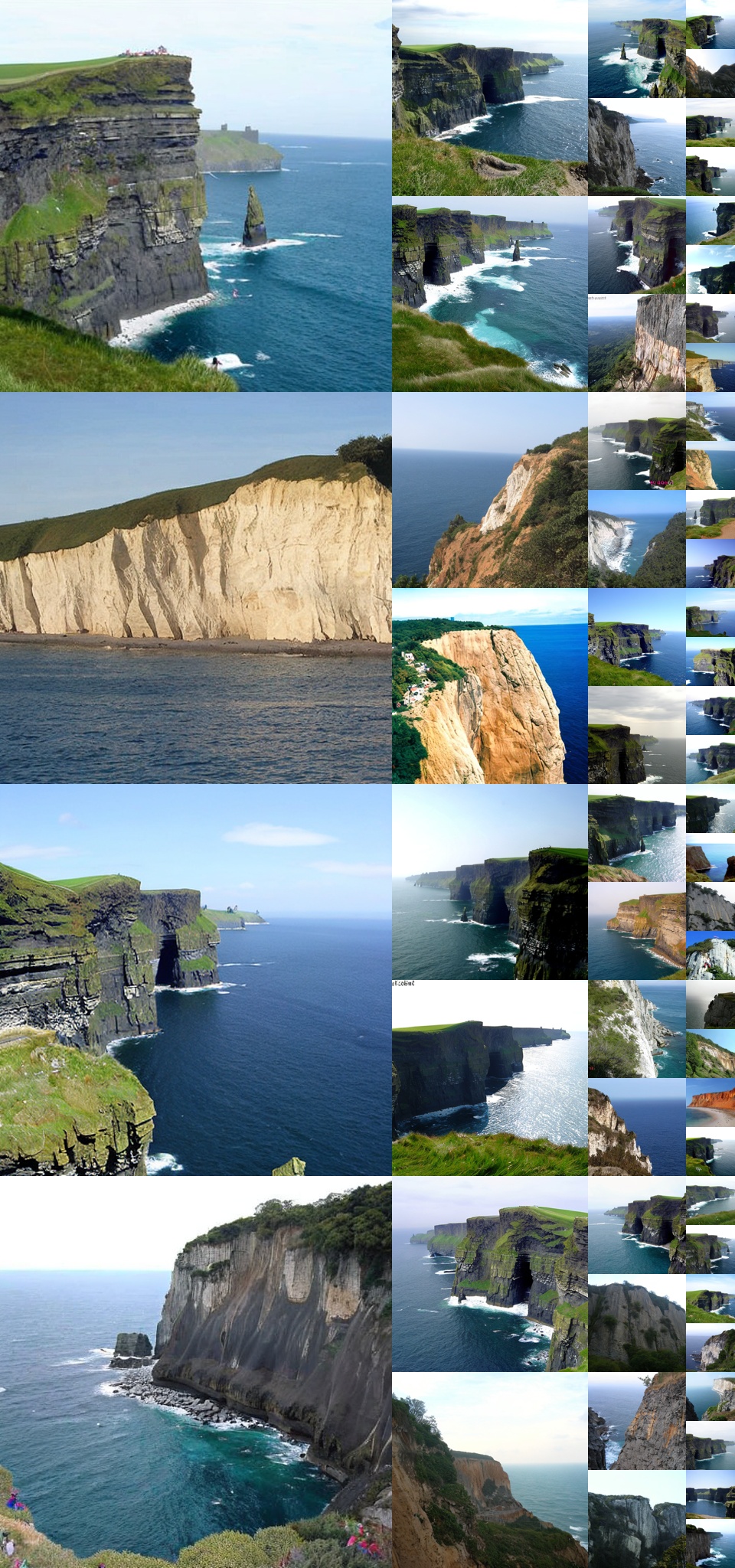}
\caption{\textbf{Uncurated $512\times512$ CausalFusion-XL samples.} \\Classifier-free guidance scale = 4.0\\Class label = ``cliff" (972)}\vspace{-2mm}
\label{fig:samples512_4}
\end{figure}

\begin{figure}\centering
\includegraphics[width=\linewidth]{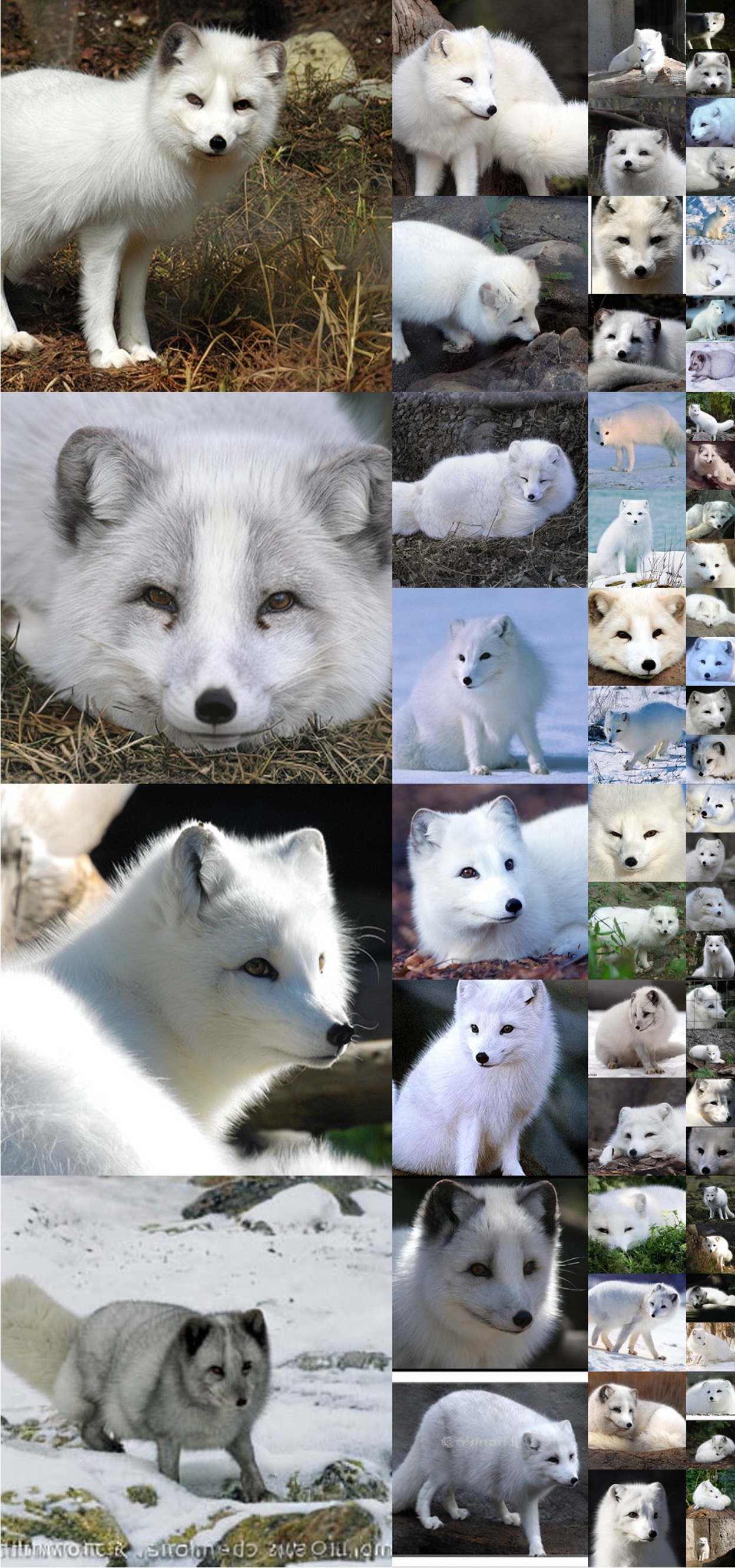}
\caption{\textbf{Uncurated $512\times512$ CausalFusion-XL samples.} \\Classifier-free guidance scale = 4.0\\Class label = ``arctic fox" (279)}\vspace{-2mm}
\label{fig:samples512_5}
\end{figure}

\begin{figure}\centering
\includegraphics[width=\linewidth]{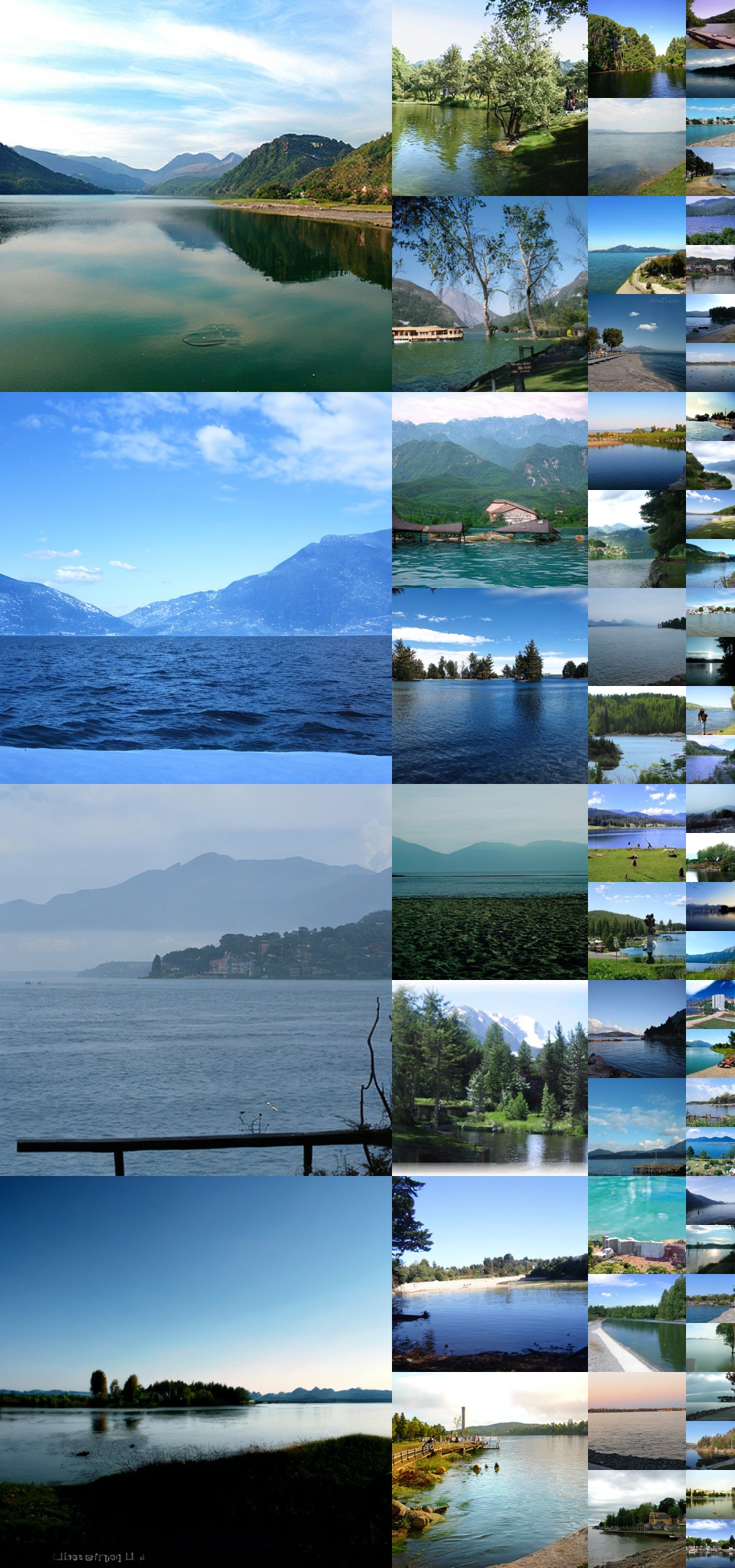}
\caption{\textbf{Uncurated $512\times512$ CausalFusion-XL samples.} \\Classifier-free guidance scale = 4.0\\Class label = ``lakeshore" (975)}\vspace{-2mm}
\label{fig:samples512_6}
\end{figure}

\begin{figure}\centering
\includegraphics[width=\linewidth]{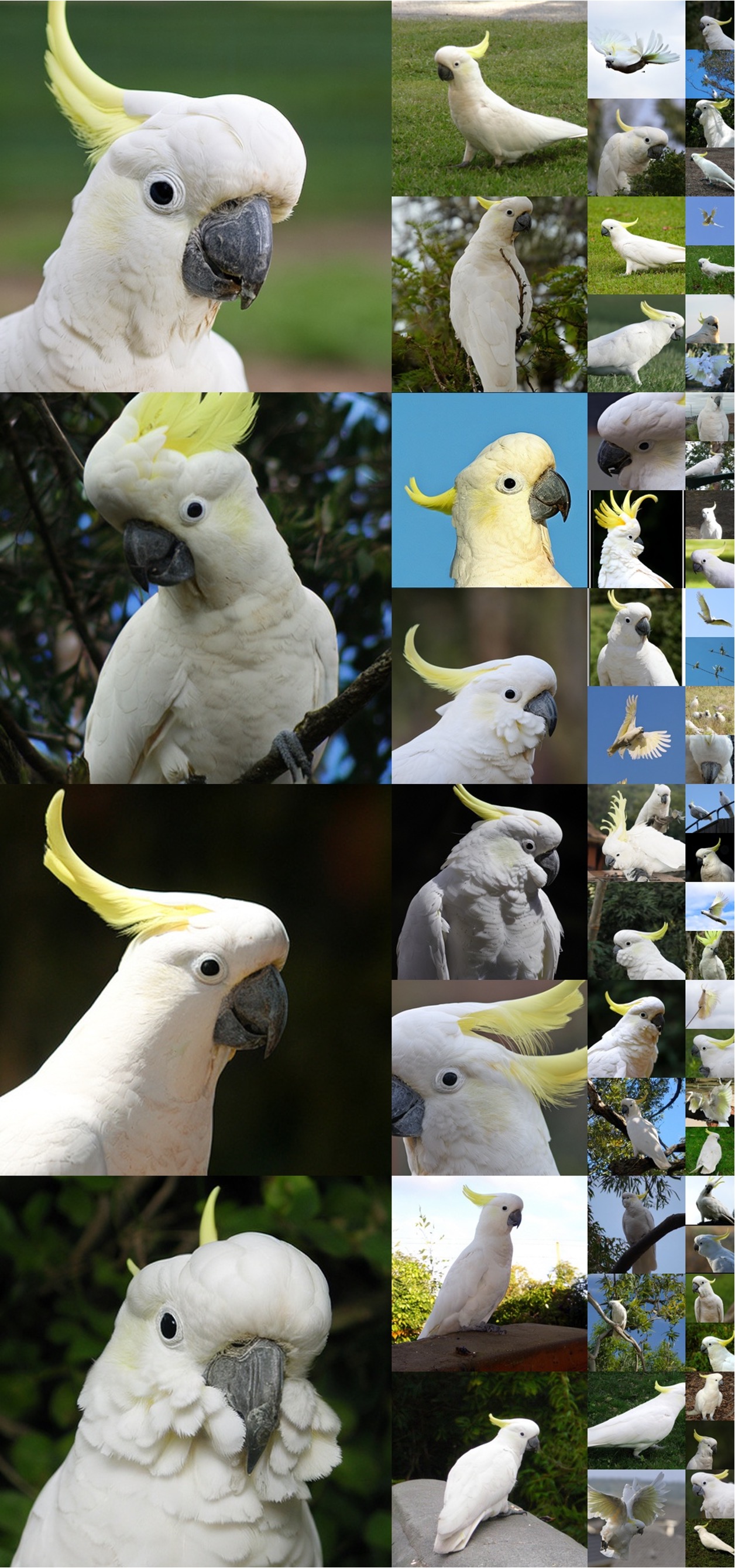}
\caption{\textbf{Uncurated $512\times512$ CausalFusion-XL samples.} \\Classifier-free guidance scale = 4.0\\Class label = ``sulphur-crested cockatoo" (89)}\vspace{-2mm}
\label{fig:samples512_7}
\end{figure}

\begin{figure}\centering
\includegraphics[width=\linewidth]{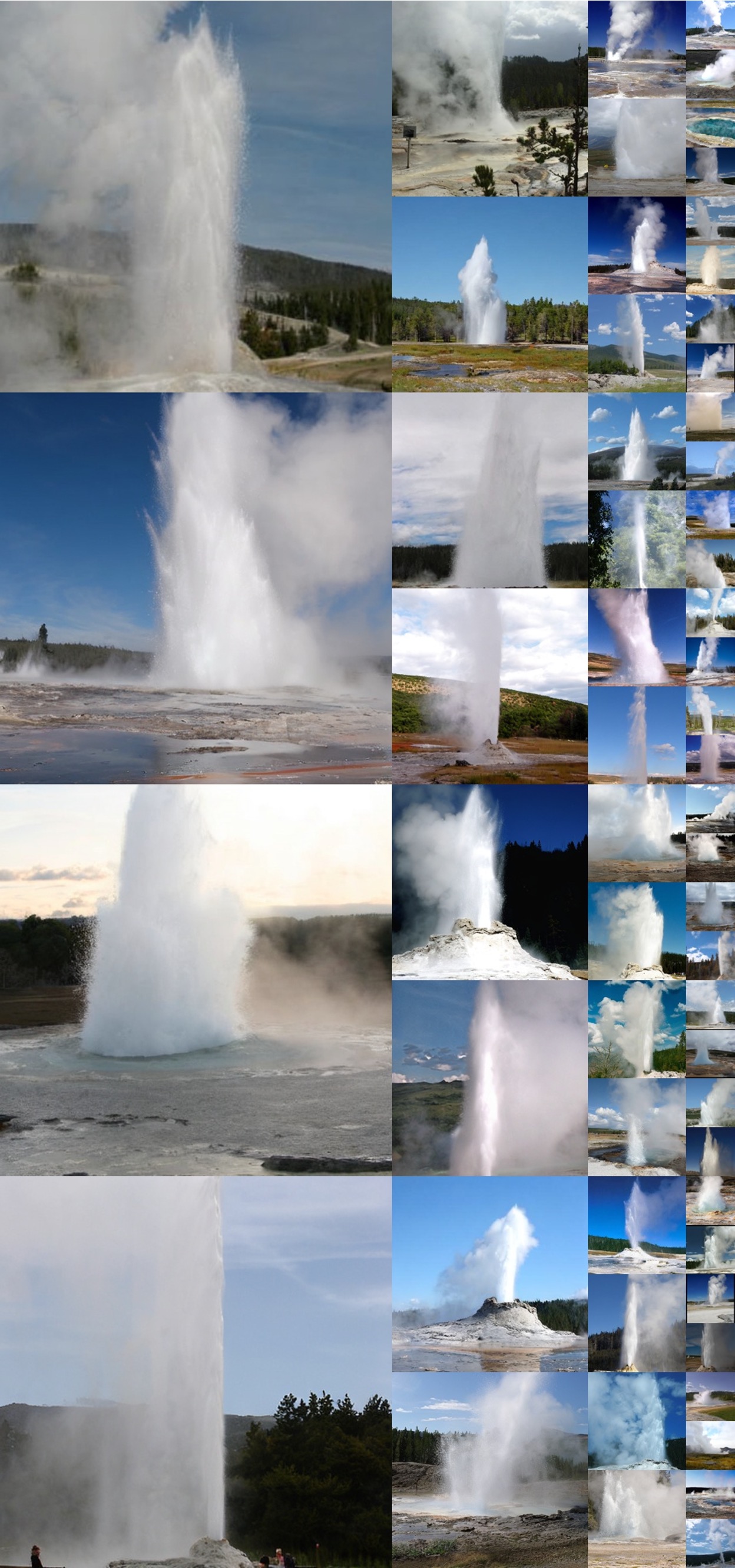}
\caption{\textbf{Uncurated $512\times512$ CausalFusion-XL samples.} \\Classifier-free guidance scale = 4.0\\Class label = ``geyser" (974)}\vspace{-2mm}
\label{fig:samples512_8}
\end{figure}

%% file: main.bbl
\begin{thebibliography}{75}
\providecommand{\natexlab}[1]{#1}
\providecommand{\url}[1]{\texttt{#1}}
\expandafter\ifx\csname urlstyle\endcsname\relax
  \providecommand{\doi}[1]{doi: #1}\else
  \providecommand{\doi}{doi: \begingroup \urlstyle{rm}\Url}\fi

\bibitem[Bao et~al.(2023)Bao, Nie, Xue, Cao, Li, Su, and Zhu]{bao2023all}
Fan Bao, Shen Nie, Kaiwen Xue, Yue Cao, Chongxuan Li, Hang Su, and Jun Zhu.
\newblock All are worth words: A vit backbone for diffusion models.
\newblock In \emph{Proceedings of the IEEE/CVF conference on computer vision and pattern recognition}, pages 22669--22679, 2023.

\bibitem[Bao et~al.(2021)Bao, Dong, Piao, and Wei]{beit}
Hangbo Bao, Li Dong, Songhao Piao, and Furu Wei.
\newblock Beit: Bert pre-training of image transformers.
\newblock In \emph{International Conference on Learning Representations}, 2021.

\bibitem[Bengio et~al.(2015)Bengio, Vinyals, Jaitly, and Shazeer]{bengio2015scheduled}
Samy Bengio, Oriol Vinyals, Navdeep Jaitly, and Noam Shazeer.
\newblock Scheduled sampling for sequence prediction with recurrent neural networks.
\newblock \emph{Advances in neural information processing systems}, 28, 2015.

\bibitem[Brooks et~al.(2024)Brooks, Peebles, Holmes, DePue, Guo, Jing, Schnurr, Taylor, Luhman, and Luhman]{sora}
Tim Brooks, Bill Peebles, Connor Holmes, Will DePue, Yufei Guo, Li Jing, David Schnurr, Joe Taylor, Troy Luhman, and Eric Luhman.
\newblock Video generation models as world simulators.
\newblock \emph{OpenAI Blog}, 2024.

\bibitem[Brown et~al.(2020)Brown, Mann, Ryder, Subbiah, Kaplan, Dhariwal, Neelakantan, Shyam, Sastry, Askell, Agarwal, Herbert-Voss, Krueger, Henighan, Child, Ramesh, Ziegler, Wu, Winter, Hesse, Chen, Sigler, Litwin, Gray, Chess, Clark, Berner, McCandlish, Radford, Sutskever, and Amodei]{gpt3}
Tom Brown, Benjamin Mann, Nick Ryder, Melanie Subbiah, Jared~D Kaplan, Prafulla Dhariwal, Arvind Neelakantan, Pranav Shyam, Girish Sastry, Amanda Askell, Sandhini Agarwal, Ariel Herbert-Voss, Gretchen Krueger, Tom Henighan, Rewon Child, Aditya Ramesh, Daniel Ziegler, Jeffrey Wu, Clemens Winter, Chris Hesse, Mark Chen, Eric Sigler, Mateusz Litwin, Scott Gray, Benjamin Chess, Jack Clark, Christopher Berner, Sam McCandlish, Alec Radford, Ilya Sutskever, and Dario Amodei.
\newblock Language models are few-shot learners.
\newblock \emph{Advances in neural information processing systems}, 33:\penalty0 1877--1901, 2020.

\bibitem[Chang et~al.(2022)Chang, Zhang, Jiang, Liu, and Freeman]{chang2022maskgit}
Huiwen Chang, Han Zhang, Lu Jiang, Ce Liu, and William~T Freeman.
\newblock Maskgit: Masked generative image transformer.
\newblock In \emph{Proceedings of the IEEE/CVF Conference on Computer Vision and Pattern Recognition}, pages 11315--11325, 2022.

\bibitem[Chang et~al.(2023)Chang, Zhang, Barber, Maschinot, Lezama, Jiang, Yang, Murphy, Freeman, Rubinstein, et~al.]{muse}
Huiwen Chang, Han Zhang, Jarred Barber, AJ Maschinot, Jos{\'e} Lezama, Lu Jiang, Ming-Hsuan Yang, Kevin Murphy, William~T Freeman, Michael Rubinstein, et~al.
\newblock Muse: Text-to-image generation via masked generative transformers.
\newblock In \emph{Proceedings of the 40th International Conference on Machine Learning}, pages 4055--4075, 2023.

\bibitem[Chen et~al.(2024)Chen, Ge, Xie, Wu, Yao, Ren, Wang, Luo, Lu, and Li]{chen2024pixart}
Junsong Chen, Chongjian Ge, Enze Xie, Yue Wu, Lewei Yao, Xiaozhe Ren, Zhongdao Wang, Ping Luo, Huchuan Lu, and Zhenguo Li.
\newblock Pixart-$\backslash$sigma: Weak-to-strong training of diffusion transformer for 4k text-to-image generation.
\newblock \emph{arXiv preprint arXiv:2403.04692}, 2024.

\bibitem[Clark et~al.(2020)Clark, Luong, Le, and Manning]{electra}
Kevin Clark, Minh-Thang Luong, Quoc~V. Le, and Christopher~D. Manning.
\newblock Electra: Pre-training text encoders as discriminators rather than generators.
\newblock In \emph{International Conference on Learning Representations}, 2020.

\bibitem[Cubuk et~al.(2020)Cubuk, Zoph, Shlens, and Le]{randaugment}
Ekin~D Cubuk, Barret Zoph, Jonathon Shlens, and Quoc~V Le.
\newblock Randaugment: Practical automated data augmentation with a reduced search space.
\newblock In \emph{Proceedings of the IEEE/CVF conference on computer vision and pattern recognition workshops}, pages 702--703, 2020.

\bibitem[Dehghani et~al.(2023)Dehghani, Djolonga, Mustafa, Padlewski, Heek, Gilmer, Steiner, Caron, Geirhos, Alabdulmohsin, et~al.]{dehghani2023vit22b}
Mostafa Dehghani, Josip Djolonga, Basil Mustafa, Piotr Padlewski, Jonathan Heek, Justin Gilmer, Andreas~Peter Steiner, Mathilde Caron, Robert Geirhos, Ibrahim Alabdulmohsin, et~al.
\newblock Scaling vision transformers to 22 billion parameters.
\newblock In \emph{International Conference on Machine Learning}, pages 7480--7512. PMLR, 2023.

\bibitem[Deng et~al.(2009)Deng, Dong, Socher, Li, Li, and Fei-Fei]{deng2009imagenet}
Jia Deng, Wei Dong, Richard Socher, Li-Jia Li, Kai Li, and Li Fei-Fei.
\newblock Imagenet: A large-scale hierarchical image database.
\newblock In \emph{2009 IEEE conference on computer vision and pattern recognition}, pages 248--255. Ieee, 2009.

\bibitem[Dhariwal and Nichol(2021)]{adm}
Prafulla Dhariwal and Alexander Nichol.
\newblock Diffusion models beat gans on image synthesis.
\newblock \emph{Advances in neural information processing systems}, 34:\penalty0 8780--8794, 2021.

\bibitem[Ding et~al.(2021)Ding, Yang, Hong, Zheng, Zhou, Yin, Lin, Zou, Shao, Yang, and Tang]{ding2021cogview}
Ming Ding, Zhuoyi Yang, Wenyi Hong, Wendi Zheng, Chang Zhou, Da Yin, Junyang Lin, Xu Zou, Zhou Shao, Hongxia Yang, and Jie Tang.
\newblock Cogview: Mastering text-to-image generation via transformers.
\newblock \emph{Advances in neural information processing systems}, 34:\penalty0 19822--19835, 2021.

\bibitem[Dosovitskiy et~al.(2021)Dosovitskiy, Beyer, Kolesnikov, Weissenborn, Zhai, Unterthiner, Dehghani, Minderer, Heigold, Gelly, Uszkoreit, and Houlsby]{vit}
Alexey Dosovitskiy, Lucas Beyer, Alexander Kolesnikov, Dirk Weissenborn, Xiaohua Zhai, Thomas Unterthiner, Mostafa Dehghani, Matthias Minderer, Georg Heigold, Sylvain Gelly, Jakob Uszkoreit, and Neil Houlsby.
\newblock An image is worth 16x16 words: Transformers for image recognition at scale.
\newblock In \emph{International Conference on Learning Representations}, 2021.

\bibitem[Dubey et~al.(2024)Dubey, Jauhri, Pandey, Kadian, Al-Dahle, Letman, Mathur, Schelten, Yang, Fan, et~al.]{llama3}
Abhimanyu Dubey, Abhinav Jauhri, Abhinav Pandey, Abhishek Kadian, Ahmad Al-Dahle, Aiesha Letman, Akhil Mathur, Alan Schelten, Amy Yang, Angela Fan, et~al.
\newblock The llama 3 herd of models.
\newblock \emph{arXiv:2407.21783}, 2024.

\bibitem[Esser et~al.(2024)Esser, Kulal, Blattmann, Entezari, M{\"u}ller, Saini, Levi, Lorenz, Sauer, Boesel, et~al.]{li2023scaling}
Patrick Esser, Sumith Kulal, Andreas Blattmann, Rahim Entezari, Jonas M{\"u}ller, Harry Saini, Yam Levi, Dominik Lorenz, Axel Sauer, Frederic Boesel, et~al.
\newblock Scaling rectified flow transformers for high-resolution image synthesis.
\newblock In \emph{Forty-first International Conference on Machine Learning}, 2024.

\bibitem[Fei et~al.(2024)Fei, Fan, Yu, Li, and Huang]{fei2024scaling}
Zhengcong Fei, Mingyuan Fan, Changqian Yu, Debang Li, and Junshi Huang.
\newblock Scaling diffusion transformers to 16 billion parameters.
\newblock \emph{arXiv preprint arXiv:2407.11633}, 2024.

\bibitem[Gafni et~al.(2022)Gafni, Polyak, Ashual, Sheynin, Parikh, and Taigman]{gafni2022make}
Oran Gafni, Adam Polyak, Oron Ashual, Shelly Sheynin, Devi Parikh, and Yaniv Taigman.
\newblock Make-a-scene: Scene-based text-to-image generation with human priors.
\newblock In \emph{European Conference on Computer Vision}, pages 89--106. Springer, 2022.

\bibitem[Gao et~al.(2024)Gao, Zhuo, Lin, Liu, Chen, Du, Xie, Luo, Qiu, Zhang, et~al.]{gao2024lumina}
Peng Gao, Le Zhuo, Ziyi Lin, Chris Liu, Junsong Chen, Ruoyi Du, Enze Xie, Xu Luo, Longtian Qiu, Yuhang Zhang, et~al.
\newblock Lumina-t2x: Transforming text into any modality, resolution, and duration via flow-based large diffusion transformers.
\newblock \emph{arXiv preprint arXiv:2405.05945}, 2024.

\bibitem[Gu et~al.(2024)Gu, Wang, Zhang, Zhang, Zhang, Jaitly, Susskind, and Zhai]{dart}
Jiatao Gu, Yuyang Wang, Yizhe Zhang, Qihang Zhang, Dinghuai Zhang, Navdeep Jaitly, Josh Susskind, and Shuangfei Zhai.
\newblock Dart: Denoising autoregressive transformer for scalable text-to-image generation.
\newblock \emph{arXiv preprint arXiv:2410.08159}, 2024.

\bibitem[Hang et~al.(2023)Hang, Gu, Li, Bao, Chen, Hu, Geng, and Guo]{minsnr}
Tiankai Hang, Shuyang Gu, Chen Li, Jianmin Bao, Dong Chen, Han Hu, Xin Geng, and Baining Guo.
\newblock Efficient diffusion training via min-snr weighting strategy.
\newblock In \emph{Proceedings of the IEEE/CVF International Conference on Computer Vision}, pages 7441--7451, 2023.

\bibitem[Hao et~al.(2024)Hao, Liu, Qi, Zhao, Zi, Xiao, Han, and Wong]{bigr}
Shaozhe Hao, Xuantong Liu, Xianbiao Qi, Shihao Zhao, Bojia Zi, Rong Xiao, Kai Han, and Kwan-Yee~K Wong.
\newblock Bigr: Harnessing binary latent codes for image generation and improved visual representation capabilities.
\newblock \emph{arXiv preprint arXiv:2410.14672}, 2024.

\bibitem[He et~al.(2022)He, Chen, Xie, Li, Doll{\'a}r, and Girshick]{he2022masked}
Kaiming He, Xinlei Chen, Saining Xie, Yanghao Li, Piotr Doll{\'a}r, and Ross Girshick.
\newblock Masked autoencoders are scalable vision learners.
\newblock In \emph{Proceedings of the IEEE/CVF conference on computer vision and pattern recognition}, pages 16000--16009, 2022.

\bibitem[Ho and Salimans(2022)]{ho2022classifier}
Jonathan Ho and Tim Salimans.
\newblock Classifier-free diffusion guidance.
\newblock \emph{arXiv preprint arXiv:2207.12598}, 2022.

\bibitem[Ho et~al.(2020)Ho, Jain, and Abbeel]{ddpm}
Jonathan Ho, Ajay Jain, and Pieter Abbeel.
\newblock Denoising diffusion probabilistic models.
\newblock \emph{Advances in neural information processing systems}, 33:\penalty0 6840--6851, 2020.

\bibitem[Hoogeboom et~al.(2023)Hoogeboom, Heek, and Salimans]{hoogeboom2023simple}
Emiel Hoogeboom, Jonathan Heek, and Tim Salimans.
\newblock simple diffusion: End-to-end diffusion for high resolution images.
\newblock In \emph{International Conference on Machine Learning}, pages 13213--13232. PMLR, 2023.

\bibitem[Huang et~al.(2016)Huang, Sun, Liu, Sedra, and Weinberger]{droppath}
Gao Huang, Yu Sun, Zhuang Liu, Daniel Sedra, and Kilian~Q Weinberger.
\newblock Deep networks with stochastic depth.
\newblock In \emph{Computer Vision--ECCV 2016: 14th European Conference, Amsterdam, The Netherlands, October 11--14, 2016, Proceedings, Part IV 14}, pages 646--661. Springer, 2016.

\bibitem[Karras et~al.(2022)Karras, Aittala, Aila, and Laine]{edm}
Tero Karras, Miika Aittala, Timo Aila, and Samuli Laine.
\newblock Elucidating the design space of diffusion-based generative models.
\newblock \emph{Advances in neural information processing systems}, 35:\penalty0 26565--26577, 2022.

\bibitem[Kingma and Gao(2024)]{kingma2024understanding}
Diederik Kingma and Ruiqi Gao.
\newblock Understanding diffusion objectives as the elbo with simple data augmentation.
\newblock \emph{Advances in Neural Information Processing Systems}, 36, 2024.

\bibitem[Kondratyuk et~al.(2024)Kondratyuk, Yu, Gu, Lezama, Huang, Schindler, Hornung, Birodkar, Yan, Chiu, et~al.]{kondratyukvideopoet}
Dan Kondratyuk, Lijun Yu, Xiuye Gu, Jose Lezama, Jonathan Huang, Grant Schindler, Rachel Hornung, Vighnesh Birodkar, Jimmy Yan, Ming-Chang Chiu, et~al.
\newblock Videopoet: A large language model for zero-shot video generation.
\newblock In \emph{Forty-first International Conference on Machine Learning}, 2024.

\bibitem[Kynk{\"a}{\"a}nniemi et~al.(2024)Kynk{\"a}{\"a}nniemi, Aittala, Karras, Laine, Aila, and Lehtinen]{kynkaanniemi2024applying}
Tuomas Kynk{\"a}{\"a}nniemi, Miika Aittala, Tero Karras, Samuli Laine, Timo Aila, and Jaakko Lehtinen.
\newblock Applying guidance in a limited interval improves sample and distribution quality in diffusion models.
\newblock \emph{arXiv preprint arXiv:2404.07724}, 2024.

\bibitem[Labs(2024)]{flux}
Black~Forest Labs.
\newblock Flux, 2024.

\bibitem[Li et~al.(2024)Li, Tian, Li, Deng, and He]{mar}
Tianhong Li, Yonglong Tian, He Li, Mingyang Deng, and Kaiming He.
\newblock Autoregressive image generation without vector quantization.
\newblock \emph{arXiv preprint arXiv:2406.11838}, 2024.

\bibitem[Li et~al.(2023)Li, Fan, Hu, Feichtenhofer, and He]{flip}
Yanghao Li, Haoqi Fan, Ronghang Hu, Christoph Feichtenhofer, and Kaiming He.
\newblock Scaling language-image pre-training via masking.
\newblock In \emph{Proceedings of the IEEE/CVF Conference on Computer Vision and Pattern Recognition}, pages 23390--23400, 2023.

\bibitem[Lin et~al.(2014)Lin, Maire, Belongie, Hays, Perona, Ramanan, Doll{\'a}r, and Zitnick]{lin2014microsoft}
Tsung-Yi Lin, Michael Maire, Serge Belongie, James Hays, Pietro Perona, Deva Ramanan, Piotr Doll{\'a}r, and C~Lawrence Zitnick.
\newblock Microsoft coco: Common objects in context.
\newblock In \emph{Computer Vision--ECCV 2014: 13th European Conference, Zurich, Switzerland, September 6-12, 2014, Proceedings, Part V 13}, pages 740--755. Springer, 2014.

\bibitem[Lipman et~al.(2023)Lipman, Chen, Ben-Hamu, Nickel, and Le]{lipman2023flow}
Yaron Lipman, Ricky~TQ Chen, Heli Ben-Hamu, Maximilian Nickel, and Matthew Le.
\newblock Flow matching for generative modeling.
\newblock In \emph{The Eleventh International Conference on Learning Representations}, 2023.

\bibitem[Liu et~al.(2024)Liu, Zeng, He, Yu, Shen, and Chen]{liu2024alleviating}
Qihao Liu, Zhanpeng Zeng, Ju He, Qihang Yu, Xiaohui Shen, and Liang-Chieh Chen.
\newblock Alleviating distortion in image generation via multi-resolution diffusion models.
\newblock \emph{arXiv preprint arXiv:2406.09416}, 2024.

\bibitem[Loshchilov(2017)]{loshchilov2017decoupled}
I Loshchilov.
\newblock Decoupled weight decay regularization.
\newblock \emph{arXiv preprint arXiv:1711.05101}, 2017.

\bibitem[Lu et~al.(2024)Lu, Clark, Lee, Zhang, Khosla, Marten, Hoiem, and Kembhavi]{io2}
Jiasen Lu, Christopher Clark, Sangho Lee, Zichen Zhang, Savya Khosla, Ryan Marten, Derek Hoiem, and Aniruddha Kembhavi.
\newblock Unified-io 2: Scaling autoregressive multimodal models with vision language audio and action.
\newblock In \emph{Proceedings of the IEEE/CVF Conference on Computer Vision and Pattern Recognition}, pages 26439--26455, 2024.

\bibitem[Luo et~al.(2024)Luo, Shi, Ge, Yang, Wang, and Shan]{luo2024open}
Zhuoyan Luo, Fengyuan Shi, Yixiao Ge, Yujiu Yang, Limin Wang, and Ying Shan.
\newblock Open-magvit2: An open-source project toward democratizing auto-regressive visual generation.
\newblock \emph{arXiv preprint arXiv:2409.04410}, 2024.

\bibitem[Ma et~al.(2024)Ma, Goldstein, Albergo, Boffi, Vanden-Eijnden, and Xie]{sit}
Nanye Ma, Mark Goldstein, Michael~S Albergo, Nicholas~M Boffi, Eric Vanden-Eijnden, and Saining Xie.
\newblock Sit: Exploring flow and diffusion-based generative models with scalable interpolant transformers.
\newblock \emph{arXiv preprint arXiv:2401.08740}, 2024.

\bibitem[Nichol et~al.(2021)Nichol, Dhariwal, Ramesh, Shyam, Mishkin, McGrew, Sutskever, and Chen]{nichol2021glide}
Alex Nichol, Prafulla Dhariwal, Aditya Ramesh, Pranav Shyam, Pamela Mishkin, Bob McGrew, Ilya Sutskever, and Mark Chen.
\newblock Glide: Towards photorealistic image generation and editing with text-guided diffusion models.
\newblock \emph{arXiv preprint arXiv:2112.10741}, 2021.

\bibitem[Peebles and Xie(2023)]{dit}
William Peebles and Saining Xie.
\newblock Scalable diffusion models with transformers.
\newblock In \emph{Proceedings of the IEEE/CVF International Conference on Computer Vision}, pages 4195--4205, 2023.

\bibitem[Podell et~al.(2024)Podell, English, Lacey, Blattmann, Dockhorn, M{\"u}ller, Penna, and Rombach]{sdxl}
Dustin Podell, Zion English, Kyle Lacey, Andreas Blattmann, Tim Dockhorn, Jonas M{\"u}ller, Joe Penna, and Robin Rombach.
\newblock Sdxl: Improving latent diffusion models for high-resolution image synthesis.
\newblock In \emph{The Twelfth International Conference on Learning Representations}, 2024.

\bibitem[Radford and Narasimhan(2018)]{gpt1}
Alec Radford and Karthik Narasimhan.
\newblock Improving language understanding by generative pre-training.
\newblock \emph{OpenAI blog}, 2018.

\bibitem[Radford et~al.(2019)Radford, Wu, Child, Luan, Amodei, Sutskever, et~al.]{gpt2}
Alec Radford, Jeffrey Wu, Rewon Child, David Luan, Dario Amodei, Ilya Sutskever, et~al.
\newblock Language models are unsupervised multitask learners.
\newblock \emph{OpenAI blog}, 2019.

\bibitem[Ramesh et~al.(2021)Ramesh, Pavlov, Goh, Gray, Voss, Radford, Chen, and Sutskever]{dalle1}
Aditya Ramesh, Mikhail Pavlov, Gabriel Goh, Scott Gray, Chelsea Voss, Alec Radford, Mark Chen, and Ilya Sutskever.
\newblock Zero-shot text-to-image generation.
\newblock In \emph{International conference on machine learning}, pages 8821--8831. Pmlr, 2021.

\bibitem[Ramesh et~al.(2022)Ramesh, Dhariwal, Nichol, Chu, and Chen]{dalle2}
Aditya Ramesh, Prafulla Dhariwal, Alex Nichol, Casey Chu, and Mark Chen.
\newblock Hierarchical text-conditional image generation with clip latents.
\newblock \emph{arXiv:2204.06125}, 2022.

\bibitem[Rombach et~al.(2022)Rombach, Blattmann, Lorenz, Esser, and Ommer]{rombach2022high}
Robin Rombach, Andreas Blattmann, Dominik Lorenz, Patrick Esser, and Bj{\"o}rn Ommer.
\newblock High-resolution image synthesis with latent diffusion models.
\newblock In \emph{Proceedings of the IEEE/CVF conference on computer vision and pattern recognition}, pages 10684--10695, 2022.

\bibitem[Sohl-Dickstein et~al.(2015{\natexlab{a}})Sohl-Dickstein, Weiss, Maheswaranathan, and Ganguli]{dpm}
Jascha Sohl-Dickstein, Eric Weiss, Niru Maheswaranathan, and Surya Ganguli.
\newblock Deep unsupervised learning using nonequilibrium thermodynamics.
\newblock In \emph{International conference on machine learning}, pages 2256--2265. PMLR, 2015{\natexlab{a}}.

\bibitem[Sohl-Dickstein et~al.(2015{\natexlab{b}})Sohl-Dickstein, Weiss, Maheswaranathan, and Ganguli]{sohl2015deep}
Jascha Sohl-Dickstein, Eric Weiss, Niru Maheswaranathan, and Surya Ganguli.
\newblock Deep unsupervised learning using nonequilibrium thermodynamics.
\newblock In \emph{International conference on machine learning}, pages 2256--2265. PMLR, 2015{\natexlab{b}}.

\bibitem[Song and Ermon(2019)]{song0}
Yang Song and Stefano Ermon.
\newblock Generative modeling by estimating gradients of the data distribution.
\newblock \emph{Advances in neural information processing systems}, 32, 2019.

\bibitem[Song et~al.(2021)Song, Sohl-Dickstein, Kingma, Kumar, Ermon, and Poole]{songscore}
Yang Song, Jascha Sohl-Dickstein, Diederik~P Kingma, Abhishek Kumar, Stefano Ermon, and Ben Poole.
\newblock Score-based generative modeling through stochastic differential equations.
\newblock In \emph{International Conference on Learning Representations}, 2021.

\bibitem[StabilityAI(2024)]{SD-vae}
StabilityAI.
\newblock \url{https://huggingface.co/stabilityai/sd-vae-ft-ema}, 2024.

\bibitem[Sun et~al.(2024{\natexlab{a}})Sun, Jiang, Chen, Zhang, Peng, Luo, and Yuan]{sun2024autoregressive}
Peize Sun, Yi Jiang, Shoufa Chen, Shilong Zhang, Bingyue Peng, Ping Luo, and Zehuan Yuan.
\newblock Autoregressive model beats diffusion: Llama for scalable image generation.
\newblock \emph{arXiv preprint arXiv:2406.06525}, 2024{\natexlab{a}}.

\bibitem[Sun et~al.(2024{\natexlab{b}})Sun, Cui, Zhang, Zhang, Yu, Wang, Rao, Liu, Huang, and Wang]{emu2}
Quan Sun, Yufeng Cui, Xiaosong Zhang, Fan Zhang, Qiying Yu, Yueze Wang, Yongming Rao, Jingjing Liu, Tiejun Huang, and Xinlong Wang.
\newblock Generative multimodal models are in-context learners.
\newblock In \emph{Proceedings of the IEEE/CVF Conference on Computer Vision and Pattern Recognition}, pages 14398--14409, 2024{\natexlab{b}}.

\bibitem[Szegedy et~al.(2016)Szegedy, Vanhoucke, Ioffe, Shlens, and Wojna]{label_smooth}
Christian Szegedy, Vincent Vanhoucke, Sergey Ioffe, Jon Shlens, and Zbigniew Wojna.
\newblock Rethinking the inception architecture for computer vision.
\newblock In \emph{Proceedings of the IEEE conference on computer vision and pattern recognition}, pages 2818--2826, 2016.

\bibitem[Teng et~al.(2023)Teng, Zheng, Ding, Hong, Wangni, Yang, and Tang]{teng2023relay}
Jiayan Teng, Wendi Zheng, Ming Ding, Wenyi Hong, Jianqiao Wangni, Zhuoyi Yang, and Jie Tang.
\newblock Relay diffusion: Unifying diffusion process across resolutions for image synthesis.
\newblock \emph{arXiv preprint arXiv:2309.03350}, 2023.

\bibitem[Tian et~al.(2024)Tian, Jiang, Yuan, Peng, and Wang]{tian2024visual}
Keyu Tian, Yi Jiang, Zehuan Yuan, Bingyue Peng, and Liwei Wang.
\newblock Visual autoregressive modeling: Scalable image generation via next-scale prediction.
\newblock \emph{arXiv preprint arXiv:2404.02905}, 2024.

\bibitem[Touvron et~al.(2023{\natexlab{a}})Touvron, Lavril, Izacard, Martinet, Lachaux, Lacroix, Rozi{\`e}re, Goyal, Hambro, Azhar, et~al.]{llama1}
Hugo Touvron, Thibaut Lavril, Gautier Izacard, Xavier Martinet, Marie-Anne Lachaux, Timoth{\'e}e Lacroix, Baptiste Rozi{\`e}re, Naman Goyal, Eric Hambro, Faisal Azhar, et~al.
\newblock Llama: Open and efficient foundation language models.
\newblock \emph{arXiv:2302.13971}, 2023{\natexlab{a}}.

\bibitem[Touvron et~al.(2023{\natexlab{b}})Touvron, Martin, Stone, Albert, Almahairi, Babaei, Bashlykov, Batra, Bhargava, Bhosale, et~al.]{llama2}
Hugo Touvron, Louis Martin, Kevin Stone, Peter Albert, Amjad Almahairi, Yasmine Babaei, Nikolay Bashlykov, Soumya Batra, Prajjwal Bhargava, Shruti Bhosale, et~al.
\newblock Llama 2: Open foundation and fine-tuned chat models.
\newblock \emph{arXiv:2307.09288}, 2023{\natexlab{b}}.

\bibitem[Tschannen et~al.(2025)Tschannen, Eastwood, and Mentzer]{tschannen2025givt}
Michael Tschannen, Cian Eastwood, and Fabian Mentzer.
\newblock Givt: Generative infinite-vocabulary transformers.
\newblock In \emph{European Conference on Computer Vision}, pages 292--309. Springer, 2025.

\bibitem[Wang et~al.(2024{\natexlab{a}})Wang, Bai, Tan, Wang, Fan, Bai, Chen, Liu, Wang, Ge, Fan, Dang, Du, Ren, Men, Liu, Zhou, Zhou, and Lin]{qwen2vl}
Peng Wang, Shuai Bai, Sinan Tan, Shijie Wang, Zhihao Fan, Jinze Bai, Keqin Chen, Xuejing Liu, Jialin Wang, Wenbin Ge, Yang Fan, Kai Dang, Mengfei Du, Xuancheng Ren, Rui Men, Dayiheng Liu, Chang Zhou, Jingren Zhou, and Junyang Lin.
\newblock Qwen2-vl: Enhancing vision-language model's perception of the world at any resolution.
\newblock \emph{arXiv:2409.12191}, 2024{\natexlab{a}}.

\bibitem[Wang et~al.(2018)Wang, Girshick, Gupta, and He]{wang2018non}
Xiaolong Wang, Ross Girshick, Abhinav Gupta, and Kaiming He.
\newblock Non-local neural networks.
\newblock In \emph{Proceedings of the IEEE conference on computer vision and pattern recognition}, pages 7794--7803, 2018.

\bibitem[Wang et~al.(2024{\natexlab{b}})Wang, Lu, Huang, Zhou, Ouyang, et~al.]{wang2024fitv2}
ZiDong Wang, Zeyu Lu, Di Huang, Cai Zhou, Wanli Ouyang, et~al.
\newblock Fitv2: Scalable and improved flexible vision transformer for diffusion model.
\newblock \emph{arXiv preprint arXiv:2410.13925}, 2024{\natexlab{b}}.

\bibitem[Wei et~al.(2022)Wei, Fan, Xie, Wu, Yuille, and Feichtenhofer]{maskedpredict}
Chen Wei, Haoqi Fan, Saining Xie, Chao-Yuan Wu, Alan Yuille, and Christoph Feichtenhofer.
\newblock Masked feature prediction for self-supervised visual pre-training.
\newblock In \emph{Proceedings of the IEEE/CVF Conference on Computer Vision and Pattern Recognition}, pages 14668--14678, 2022.

\bibitem[Yu et~al.(2022)Yu, Xu, Koh, Luong, Baid, Wang, Vasudevan, Ku, Yang, Ayan, et~al.]{parti}
Jiahui Yu, Yuanzhong Xu, Jing~Yu Koh, Thang Luong, Gunjan Baid, Zirui Wang, Vijay Vasudevan, Alexander Ku, Yinfei Yang, Burcu~Karagol Ayan, et~al.
\newblock Scaling autoregressive models for content-rich text-to-image generation.
\newblock \emph{Transactions on Machine Learning Research}, 2022.

\bibitem[Yun et~al.(2019)Yun, Han, Oh, Chun, Choe, and Yoo]{cutmix}
Sangdoo Yun, Dongyoon Han, Seong~Joon Oh, Sanghyuk Chun, Junsuk Choe, and Youngjoon Yoo.
\newblock Cutmix: Regularization strategy to train strong classifiers with localizable features.
\newblock In \emph{Proceedings of the IEEE/CVF international conference on computer vision}, pages 6023--6032, 2019.

\bibitem[Zhan et~al.(2024)Zhan, Dai, Ye, Zhou, Zhang, Liu, Zhang, Yuan, Zhang, Li, et~al.]{anygpt}
Jun Zhan, Junqi Dai, Jiasheng Ye, Yunhua Zhou, Dong Zhang, Zhigeng Liu, Xin Zhang, Ruibin Yuan, Ge Zhang, Linyang Li, et~al.
\newblock Anygpt: Unified multimodal llm with discrete sequence modeling.
\newblock \emph{arXiv:2402.12226}, 2024.

\bibitem[Zhang et~al.(2018)Zhang, Cisse, Dauphin, and Lopez-Paz]{mixup}
Hongyi Zhang, Moustapha Cisse, Yann~N. Dauphin, and David Lopez-Paz.
\newblock mixup: Beyond empirical risk minimization.
\newblock In \emph{International Conference on Learning Representations}, 2018.

\bibitem[Zhao et~al.(2024)Zhao, Song, Wang, Feng, Ding, Sun, Xiao, and Wang]{monoformer}
Chuyang Zhao, Yuxing Song, Wenhao Wang, Haocheng Feng, Errui Ding, Yifan Sun, Xinyan Xiao, and Jingdong Wang.
\newblock Monoformer: One transformer for both diffusion and autoregression.
\newblock \emph{arXiv:2409.16280}, 2024.

\bibitem[Zheng et~al.(2023)Zheng, Nie, Vahdat, and Anandkumar]{zheng2023fast}
Hongkai Zheng, Weili Nie, Arash Vahdat, and Anima Anandkumar.
\newblock Fast training of diffusion models with masked transformers.
\newblock \emph{arXiv preprint arXiv:2306.09305}, 2023.

\bibitem[Zhong et~al.(2020)Zhong, Zheng, Kang, Li, and Yang]{erasing}
Zhun Zhong, Liang Zheng, Guoliang Kang, Shaozi Li, and Yi Yang.
\newblock Random erasing data augmentation.
\newblock In \emph{Proceedings of the AAAI conference on artificial intelligence}, pages 13001--13008, 2020.

\bibitem[Zhou et~al.(2024)Zhou, Yu, Babu, Tirumala, Yasunaga, Shamis, Kahn, Ma, Zettlemoyer, and Levy]{transfusion}
Chunting Zhou, Lili Yu, Arun Babu, Kushal Tirumala, Michihiro Yasunaga, Leonid Shamis, Jacob Kahn, Xuezhe Ma, Luke Zettlemoyer, and Omer Levy.
\newblock Transfusion: Predict the next token and diffuse images with one multi-modal model.
\newblock \emph{arXiv:2408.11039}, 2024.

\end{thebibliography}
